\documentclass{article}

\PassOptionsToPackage{numbers, compress}{natbib}

\usepackage[preprint]{neurips_2024}

\usepackage[utf8]{inputenc} 
\usepackage[T1]{fontenc}    
\usepackage{url}            
\usepackage{booktabs}       
\usepackage{amsfonts}       
\usepackage{nicefrac}       
\usepackage{microtype}      
\usepackage{xcolor}         
\usepackage{times}
\usepackage{epsfig}
\usepackage{amsmath}
\usepackage{makecell} 
\usepackage{tabularx}
\usepackage{diagbox}
\usepackage{enumitem} 

\usepackage{xspace}
\usepackage{graphicx}
\usepackage{subcaption}
\usepackage{multirow} 
\usepackage{colortbl}
\usepackage{pifont}
\usepackage{wrapfig}
\usepackage{tcolorbox}
\usepackage{float}
\usepackage{marvosym}
\definecolor{citecolor}{HTML}{0071BC}
\usepackage[
    pagebackref,           
    breaklinks=true,       
    colorlinks=true,       
    linkcolor=red,         
    citecolor=citecolor,   
    urlcolor=black,        
    bookmarks=false        
]{hyperref}

\newcommand{\eg}{{\emph{e.g.}}}

\newcommand{\vs}{\emph{vs.}}

\newcommand{\yes}{\ding{51}}
\newcommand{\no}{\ding{55}}

\newcommand{\gbf}[1]{\green{\bf{\fn{(#1)}}}}

\title{
InternVL3: Exploring Advanced Training and Test-Time Recipes for Open-Source Multimodal  Models
}

\author{
\scalebox{0.85}{
Jinguo Zhu$^{1*}$,
Weiyun Wang$^{5,1*\dagger}$, 
Zhe Chen$^{4,1*\dagger}$,
Zhaoyang Liu$^{1*\dagger}$,
Shenglong Ye$^{1*}$,
Lixin Gu$^{1*}$,
Hao Tian$^{2*}$,
}\\
\scalebox{0.85}{
\textbf{
Yuchen Duan$^{6,1*\dagger}$,
Weijie Su$^{1}$,
Jie Shao$^{4,1\dagger}$,
Zhangwei Gao$^{7,1\dagger}$,
Erfei Cui$^{7,1\dagger}$,
Xuehui Wang$^{7,1\dagger}$,
Yue Cao$^{4,1\dagger}$,
}}\\
\scalebox{0.85}{
\textbf{
Yangzhou Liu$^{4,1\dagger}$,
Xingguang Wei$^{1\dagger}$,
Hongjie Zhang$^{1}$,
Haomin Wang$^{7,1\dagger}$,
Weiye Xu$^{1\dagger}$,
Hao Li$^{1\dagger}$,
Jiahao Wang$^{1\dagger}$,
}}\\
\scalebox{0.85}{\textbf{
Nianchen Deng$^{1}$,
Songze Li$^{1}$,
Yinan He$^{1}$,
Tan Jiang$^{2}$,
Jiapeng Luo$^{2}$,
Yi Wang$^{1}$,
Conghui He$^{1}$,
Botian Shi$^{1}$, 
}}\\
\scalebox{0.85}{
\textbf{
Xingcheng Zhang$^{1}$,
Wenqi Shao$^{1}$,
Junjun He$^{1}$,
Yingtong Xiong$^{1}$, 
Wenwen Qu$^{1}$, 
Peng Sun$^{1}$, 
Penglong Jiao$^{1}$, 
}}\\
\scalebox{0.85}{
\textbf{
Han Lv$^{1}$,
Lijun Wu$^{1}$,
Kaipeng Zhang$^{1}$,
Huipeng Deng$^{1}$,
Jiaye Ge$^{1}$,
Kai Chen$^{1}$,
Limin Wang$^{4,1}$,
Min Dou$^{1}$,
}}\\
\scalebox{0.85}{
\textbf{
Lewei Lu$^{2}$,
Xizhou Zhu$^{3,1}$,
Tong Lu$^{4}$,
Dahua Lin$^{6,1}$,
Yu Qiao$^{1}$,
Jifeng Dai$^{3,1}$\textsuperscript{\Letter}, 
Wenhai Wang$^{6,1}$\textsuperscript{\Letter} 
}}
\vspace{4px}\\
\scalebox{0.95}{
$^1$Shanghai AI Laboratory~~~
$^2$SenseTime Research~~~
$^3$Tsinghua University~~~
$^4$Nanjing University~~~}\\
\scalebox{0.95}{
$^5$Fudan University~~~
$^6$The Chinese University of Hong Kong~~~
$^7$Shanghai Jiao Tong University
}\\
\\
\small Code: \url{https://github.com/OpenGVLab/InternVL} \\
\small Model: \url{https://huggingface.co/OpenGVLab/InternVL3-78B} \\
\small Data: \url{https://huggingface.co/datasets/OpenGVLab/InternVL-Data} \\
}

\newcommand\blfootnote[1]{%
\begingroup
\renewcommand\thefootnote{}\footnote{#1}%
\addtocounter{footnote}{-1}%
\endgroup
}

\definecolor{baselinecolor}{gray}{.9}
\definecolor{reduce-color}{RGB}{67,178,68}

\newcommand{\green}[1]{\textcolor[RGB]{96,177,87}{#1}}
\newcommand{\fn}[1]{\footnotesize{#1}}

\begin{document}

\maketitle

\blfootnote{
* equal contribution; $\dagger$ interns at OpenGVLab, Shanghai AI Laboratory; \\
\Letter\  corresponding authors (daijifeng@tsinghua.edu.cn, wangwenhai@pjlab.org.cn).
}

\begin{abstract}
 
We introduce \textit{InternVL3}, a significant advancement in the InternVL series featuring a native multimodal pre-training paradigm. Rather than adapting a text-only large language model (LLM) into a multimodal large language model (MLLM) that supports visual inputs, InternVL3 jointly acquires multimodal and linguistic capabilities from both diverse multimodal data and pure-text corpora during a single pre-training stage. 
This unified training paradigm effectively addresses the complexities and alignment challenges commonly encountered in conventional post-hoc training pipelines for MLLMs.
To further improve performance and scalability, InternVL3 incorporates variable visual position encoding (V2PE) to support extended multimodal contexts, employs advanced post-training techniques such as supervised fine-tuning (SFT) and mixed preference optimization (MPO), and adopts test-time scaling strategies alongside an optimized training infrastructure. 
Extensive empirical evaluations demonstrate that InternVL3 delivers superior performance across a wide range of multi-modal tasks. In particular, InternVL3-78B achieves a score of 72.2 on the MMMU benchmark, setting a new state-of-the-art among open-source MLLMs. Its capabilities remain highly competitive with leading proprietary models, including ChatGPT-4o, Claude 3.5 Sonnet, and Gemini 2.5 Pro, while also maintaining strong pure-language proficiency.  
In pursuit of open-science principles, we will publicly release both the training data and model weights to foster further research and development in next-generation MLLMs.
\end{abstract}

\section{Introduction}
 
Multimodal large language models (MLLMs)~\cite{dong2024xc24khd, li2021improved, wang2024qwen2vl, chen2023internvl, chen2024far, wang2023cogvlm, li2023monkey, team2023gemini, gpt4v, ye2023mplug2, lin2024vila, deitke2024molmo, lu2024bluelm, geminipro2.5, chen2024expanding, marafioti2025smolvlm, shi2024eagle, li2025eagle2} have recently achieved or even surpassed human-level performance in a broad spectrum of tasks, underscoring their potential as a significant stride toward artificial general intelligence (AGI). Yet, the majority of leading MLLMs—both open-source and proprietary—are adapted from text-only large language models through sophisticated multi-stage pipelines~\cite{chen2023internvl, chen2024far, chen2024expanding, bai2023qwenvl, wang2024qwen2vl, bai2025qwen2_5}. These ``post-hoc'' approaches are built upon the original text-based pre-training processes, thereby introducing alignment challenges when integrating additional modalities such as vision. In practice, bridging modality gaps often necessitates incorporating auxiliary data from specialized domains (e.g., optical character recognition scenarios) and intricate parameter-freezing or multi-stage fine-tuning schedules to ensure that core linguistic capacities remain uncompromised~\cite{liu2023llava,bai2025qwen2_5,bai2023qwenvl,chen2024expanding}. Such resource-intensive strategies highlight the need for more efficient multimodal training paradigms.

In this report, we introduce InternVL3, the latest milestone in the InternVL series~\cite{chen2023internvl, chen2024internvl_1_5, chen2024expanding}, which is distinguished by its native multimodal pre-training strategy. Rather than first pre-training a text-only large language model and subsequently retrofitting it via multimodal alignment to support visual processing, InternVL3 learns multimodal capabilities from the pre-training stage by jointly exposed to both text-only corpora and diverse multimodal datasets. This unified approach enables the model to simultaneously acquire linguistic and multimodal competencies in a more efficient and integrated manner.

\begin{figure}
    \centering
    \includegraphics[width=1.0\linewidth]{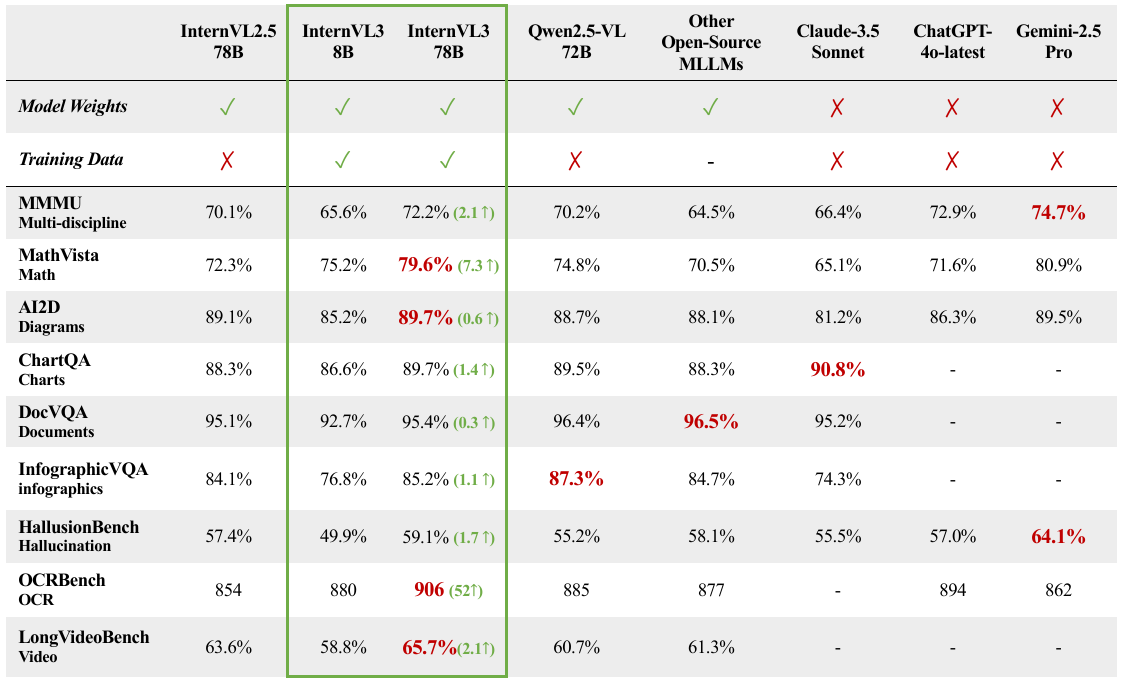}
    \caption{\textbf{Multimodal performance of the InternVL series and other advanced MLLMs.}
The InternVL series has consistently exhibited progressive enhancements in multimodal capabilities. The newly released InternVL3 significantly outperforms existing open-source MLLMs. Moreover, even in comparison with state-of-the-art closed-source commercial models, InternVL3 continues to demonstrate highly competitive performance.
}
    \label{fig:overall_performance}
    \vspace{-0.5em}
\end{figure}

InternVL3 further excels through multiple innovations that reinforce both performance and scalability. We employ a variable visual position encoding (V2PE) mechanism~\cite{ge2024v2pe} to accommodate longer multimodal contexts. Furthermore, advanced post-training strategies—comprising supervised fine-tuning (SFT) and mixed preference optimization (MPO)~\cite{wang2024mpo}—together with test-time scaling strategies~\cite{wang2025visualprm} and an optimized training infrastructure~\cite{chen2024internevo}, significantly enhance InternVL3’s efficiency and performance.

Comprehensive empirical evaluations demonstrate that InternVL3 surpasses its predecessors (\eg, InternVL2.5~\cite{chen2024expanding}) across a wide range of tasks, including multi-discipline reasoning, document understanding, multi-image / video understanding, real-world comprehension, multimodal hallucination detection, visual grounding, and multilingual capabilities. Notably, by incorporating expanded domain-specific datasets, InternVL3 also exhibits marked improvements in tool usage, GUI agents, industrial image analysis, and spatial reasoning, thus substantially extending the multimodal scenarios addressed by the InternVL series. It proves highly competitive with other open-source MLLMs such as Qwen2.5-VL~\cite{bai2025qwen2_5} and remains on par with closed-source models (\eg, ChatGPT-4o~\cite{chatgpt4o}, Claude-3.5 Sonnet~\cite{claude3series2024},  Gemini-2.5 Pro~\cite{geminipro2.5}). This versatility is evidenced by its 72.2-point performance on the MMMU benchmark~\cite{yue2023mmmu}, setting a new standard among open-source MLLMs. 
Additionally, InternVL3 demonstrates language capabilities comparable to other advanced LLMs of similar scale.

To foster further advancements within the open-source community, we will release the training data\footnote{The open-source data are  being organized, and a comprehensive list will be included in a future revision of this report.} and model weights alongside this work, thereby ensuring transparency and reproducibility for the continued development of next-generation MLLMs.

\begin{figure*}[t!]
    \centering
    \includegraphics[width=1\linewidth]{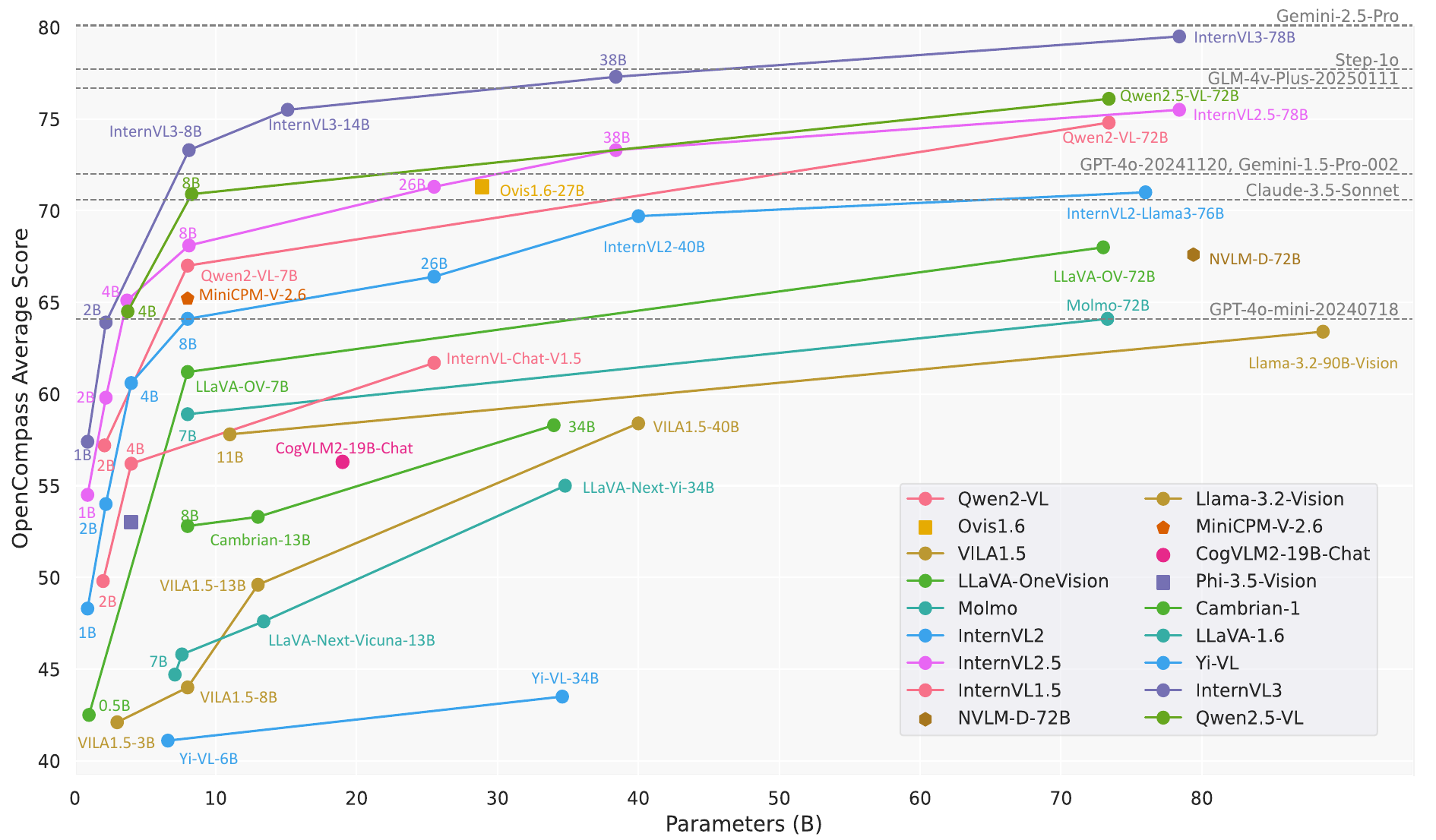}
    \caption{
    \textbf{Performance of various MLLMs on the OpenCompass multimodal academic leaderboard.} 
    The enhanced InternVL series—InternVL3—demonstrates outstanding multimodal capabilities, significantly outperforming both the Qwen2.5-VL series and closed-source models such as Step-1o, GLM-4v-Plus, and GPT-4o. Remarkably, InternVL3-78B also remains highly competitive with the state-of-the-art Gemini-2.5-Pro.
    } 
    \label{fig:oc_cruve}
\end{figure*}

\section{InternVL3}
Building upon the prior InternVL series~\cite{chen2023internvl, chen2024far, chen2024expanding}, we propose InternVL3, a new generation within the InternVL model family. InternVL3 is specifically designed to streamline the training pipeline while significantly enhancing multimodal capabilities. In this section, we first delineate the core components of InternVL3, including its model architecture, training procedures, test-time scaling strategies, and infrastructure-level optimizations.

\subsection{Model Architecture}
\label{sec:model_architecture}

The architecture of InternVL3 follows the same general framework as its predecessors, adhering to the ``ViT-MLP-LLM'' paradigm~\cite{li2021improved,chen2024expanding,gao2024mini_internvl,chen2024internvl_1_5}. Detailed architectural specifications are summarized in Table~\ref{tab:model_architecture}.

Although the native pre-training paradigm discussed later could enable training MLLMs from scratch, we choose to initialize the ViT and LLM components with pre-trained model weights to reduce computational costs. The vision encoder is available in two configurations: InternViT-300M and InternViT-6B. For the language model, we leverage pre-trained large language models (LLMs), specifically the Qwen2.5 series and InternLM3-8B. Importantly, our LLM components are initialized solely from pre-trained base models, without employing instruction-tuned variants.
The multilayer perceptron (MLP) utilized in the model is a two-layer network with random initialization. In line with the approach taken in InternVL2.5, InternVL3 incorporates a pixel unshuffle operation to enhance scalability for processing high-resolution images. This operation reduces the visual token count to one-quarter of its original value, representing each 448×448 image tile with 256 visual tokens.

\begin{table}[t]
\centering
    {\fontsize{8}{10}\selectfont 
    \renewcommand\arraystretch{1.05} 
    \setlength\tabcolsep{7pt}
    \begin{tabular}{l|c|c|c|c}
        \hline
        Model Name & \#Param & Vision Encoder & Language Model & OpenCompass Academic\\
        \hline
        \href{https://huggingface.co/OpenGVLab/InternVL3-1B}{InternVL3-1B} & 0.9B & \href{https://huggingface.co/OpenGVLab/InternViT-300M-448px-V2_5}{InternViT-300M-448px-V2.5} & \href{https://huggingface.co/Qwen/Qwen2.5-0.5B}{Qwen2.5-0.5B}  & 57.4 \\
        \href{https://huggingface.co/OpenGVLab/InternVL3-2B}{InternVL3-2B} & 1.9B & \href{https://huggingface.co/OpenGVLab/InternViT-300M-448px-V2_5}{InternViT-300M-448px-V2.5} & \href{https://huggingface.co/Qwen/Qwen2.5-1.5B}{Qwen2.5-1.5B} & 63.9\\
        \href{https://huggingface.co/OpenGVLab/InternVL3-8B}{InternVL3-8B} & 8.1B & \href{https://huggingface.co/OpenGVLab/InternViT-300M-448px-V2_5}{InternViT-300M-448px-V2.5} & \href{https://huggingface.co/Qwen/Qwen2.5-7B}{Qwen2.5-7B} & 73.3 \\
        \href{https://huggingface.co/OpenGVLab/InternVL3-9B}{InternVL3-9B} & 9.2B & \href{https://huggingface.co/OpenGVLab/InternViT-300M-448px-V2_5}{InternViT-300M-448px-V2.5} & \href{https://huggingface.co/internlm/internlm3-8b-instruct}{InternLM3-8B} & 72.4 \\
        \href{https://huggingface.co/OpenGVLab/InternVL3-14B}{InternVL3-14B} & 15.1B & \href{https://huggingface.co/OpenGVLab/InternViT-300M-448px-V2_5}{InternViT-300M-448px-V2.5} & \href{https://huggingface.co/Qwen/Qwen2.5-14B}{Qwen2.5-14B} & 75.5 \\
        \href{https://huggingface.co/OpenGVLab/InternVL3-38B}{InternVL3-38B} & 38.4B & \href{https://huggingface.co/OpenGVLab/InternViT-6B-448px-V2_5}{InternViT-6B-448px-V2.5} & \href{https://huggingface.co/Qwen/Qwen2.5-32B}{Qwen2.5-32B} & 77.3\\
        \href{https://huggingface.co/OpenGVLab/InternVL3-78B}{InternVL3-78B} & 78.4B & \href{https://huggingface.co/OpenGVLab/InternViT-6B-448px-V2_5}{InternViT-6B-448px-V2.5} & \href{https://huggingface.co/Qwen/Qwen2.5-72B}{Qwen2.5-72B} & 79.5 \\
        \hline
    \end{tabular}
    }
    \vspace{1em}
    \caption{
    \textbf{Pre-trained models used in the InternVL3 series.}
    The OpenCompass scores for the InternVL3 series were obtained through our local testing. 
    }
\label{tab:model_architecture}
\end{table}

\noindent \textbf{Variable Visual Position Encoding.} InternVL3 also integrates the \emph{Variable Visual Position Encoding} (V2PE)~\cite{ge2024v2pe}, which utilizes smaller, more flexible position increments for visual tokens. This modification facilitates the handling of longer multimodal contexts without excessively extending the position window. Specifically, each training sample for the MLLM is represented as:
\begin{equation}
    \mathbf{x} \;=\; \bigl(x_1, x_2, \dots, x_L \bigr),
\label{eq:native_x_def}
\end{equation}
where each token $x_i$ can be a textual token embedding, a visual embedding, or another modality-specific representation (\eg, video patch embeddings).  
The position index $p_i$ for any token $x_i$ can be computed sequentially as follows:
\begin{equation}
p_i = \begin{cases}
0, & \text{if } i = 1, \\
f_{\text{pos}}(p_{i-1}, x_i), & \text{for } i = 2, 3, \dots, N.
\end{cases}
\label{eq2}
\end{equation}
In contrast to traditional MLLMs, where position indices increment uniformly by 1 for each token, irrespective of modality, V2PE employs a modality-specific recursive function for position index computation. This results in distinct position index assignments for textual and visual tokens:
\begin{equation}
p_i = p_{i-1} + \begin{cases}
1, & \text{if } x_i \text{ is a textual token}, \\
\delta, & \text{if } x_i \text{ is a visual token},
\end{cases}
\label{eq:delta}
\end{equation}
where \( \delta \) is a smaller increment (\( \delta < 1 \)), reducing the rate at which position indices increase for visual tokens. The standard increment of 1 is retained for textual tokens to preserve their positional distinctions.
In line with the original V2PE design, we maintain that \( \delta \) remains constant within a single image to preserve the relative positional relationships. During training, \( \delta \) is randomly chosen for each image from a predefined set of fractional values:
\begin{equation}
\small
\delta \in \Delta = \left\{1, \frac{1}{2}, \frac{1}{4}, \frac{1}{8}, \frac{1}{16}, \frac{1}{32}, \frac{1}{64}, \frac{1}{128}, \frac{1}{256}\right\}.
\label{delta_set}
\end{equation}
During inference, \( \delta \) can be flexibly selected based on the input sequence length, enabling a balance between task performance and ensuring that position indices remain within the model's valid context range. Notably,  when \( \delta =1\), V2PE reverts to the conventional positional encoding used in InternVL2.5.

\subsection{Native Multimodal Pre-Training}
\label{sec:native_intro}

We propose a \emph{native multimodal pre-training} approach that consolidates language pre-training and multi-modal alignment training into a single pre-training stage.
Unlike conventional paradigms—where a language-only large model is first trained (typically with language pre-training followed by language post-training) and subsequently adapted to accommodate additional modalities—our method performs integrated optimization by interleaving multimodal data (e.g., image–text, video–text, or interleaved image–text sequences) with large-scale textual corpora during the pre-training process. 
This unified training scheme enables the pre-trained model to learn both linguistic and multimodal capabilities simultaneously, ultimately enhancing its capability to handle vision-language tasks without introducing additional bridging modules or subsequent inter-model alignment procedures.

\vspace{0.5em}
\noindent \textbf{Multimodal Autoregressive Formulation.}
Let $\mathcal{M}$ denote a Transformer-based model parameterized by $\theta$ that can process text, image, and video simultaneously. Specifically, for an arbitrary training sample $\mathbf{x} \;=\; \bigl(x_1, x_2, \dots, x_L \bigr)$  with the token length of $L$, we adopt the standard left-to-right autoregressive objective:
\begin{equation}
    \mathcal{L}_{\text{full}}(\theta)
    \;=\; - \sum_{i=2}^{L}
    w_{i} \cdot
    \log \, p_{\theta}\bigl(x_i \,\bigm|\; x_1,\dots, x_{i-1}\bigr),
\label{eq:autoregressive-loss-full}
\end{equation}
where $w_i$ denotes the loss weight of token $i$.
Although this formulation naturally propagates gradients through tokens of all modalities, we restrict the loss computation exclusively to \emph{text tokens}, resulting in:
\begin{equation}
    \mathcal{L}_{\text{text-only}}(\theta)
    \;=\; - \sum_{\substack{i=2 \\ x_i \,\in\, \mathrm{Text}}}^{L}
    w_{i} \cdot
    \log \, p_\theta \bigl(x_i \,\bigm|\; x_1, \dots, x_{i-1}\bigr).
\label{eq:autoregressive-loss-text}
\end{equation}
Under this selective objective, visual tokens serve as conditioning context for text prediction  and are not directly predicted.
Consequently, the model learns to embed multimodal information in a manner that is beneficial for downstream language decoding tasks. 
Notably, regarding the design choice of the token weight $w_i$, as discussed in InternVL2.5~\cite{chen2024expanding}, the widely used token averaging and sample averaging strategies can lead to gradients biased toward longer and shorter responses, respectively. To mitigate this issue, we adopt square averaging, which is defined as:

\begin{equation}
w_i =
\begin{cases}
    \frac{1}{l^0}, & \text{for token averaging} \\
    \frac{1}{l^{0.5}}, & \text{for square averaging} \\
    \frac{1}{l^1}, & \text{for sample averaging},
\end{cases}
\end{equation}

where $l$ denotes the number of tokens in the training sample on which the loss needs to be calculated.

\vspace{0.5em}
\noindent \textbf{Joint Parameter Optimization.}
Unlike the conventional ``language-only training followed by multimodal adaptation'' paradigm, our method updates \emph{all} model parameters \emph{jointly} during multimodal pre-training. Specifically, let
\begin{equation}
    \theta^*
    \;=\;\underset{\theta}{\arg \min}\;
    \mathbb{E}_{\mathbf{x}\,\in\,\mathcal{D}_{\text{multi}}}
    \bigl[\mathcal{L}_{\text{text-only}}(\theta)\bigr],
\label{eq:multimodal-argmin}
\end{equation}
where $\mathcal{D}_{\text{multi}}$ is the union of large-scale text-only and multimodal corpora (\eg, image--text or video--text pairs). We thus optimize a single model to handle these combined data sources. This multi-task joint optimization ensures that text representations and visual features are learned in concert, reinforcing alignment across modalities.

Moreover, this integrated optimization departs from conventional ``language-only training followed by multimodal adaptation'' pipelines, which often freeze or partially fine-tune certain layers in the LLM component or even in the ViT encoder when adapting to MLLM. In contrast, our method trains every layer jointly,  allowing all parameters to be jointly optimized on large-scale multimodal corpora and ensuring that both linguistic and visual features evolve synchronously. As a result, the final parameters are primed for high performance on both pure language and multimodal tasks, without additional tuning steps.

\noindent\textbf{Data.}
The pre-training data utilized in InternVL3 is broadly classified into two categories: multimodal data and pure language data. The multimodal dataset comprises a synthesis of pre-existing datasets alongside newly acquired real-world data. 
Specifically, we leverage the pre-training corpus from InternVL2.5, which covers a diverse range of domains such as image captioning, general question answering, mathematics, charts, optical character recognition (OCR), knowledge grounding, document understanding, multi-turn dialogue, and medical data. Although the overall data scale was not increased, the utility of this dataset was significantly improved by updating not only to the MLP module weights but also  to those associated with the ViT and LLM components. 
In addition, to enhance the model's ability to generalize  in real-world applications, additional data is incorporated from tasks related to graphical user interfaces (GUI), tool usage, 3D scene understanding, and video comprehension.

To compensate for the relatively short and less diverse textual content typically found in multimodal datasets, we integrate pure language data into the pre-training process. This helps preserve and amplify the model’s capabilities in language understanding and generation. 
The language corpus is primarily constructed on the pre-training data from InternLM2.5 and is further augmented with various open-source text datasets~\cite{benallal2024smollmcorpus,acemath2024,scp116k}. 
This enhancement aims to improve the model’s performance on knowledge-intensive tasks, as well as its proficiency in mathematical and reasoning tasks.

Given the complexity of balancing these heterogeneous data sources, determining an appropriate sampling strategy is non-trivial. In InternVL3, we adopt a two-stage strategy to establish the optimal sampling ratio between multimodal and language data. Initially, we train separate models on the multimodal and language datasets and evaluate their performance on corresponding benchmarks, allowing us to identify optimal sampling ratios within each modality. Then, under a fixed total training budget, we combine the two modalities and determine their relative sampling ratio. Empirical studies show that a 1:3 ratio of language to multimodal data yields the best overall performance across both unimodal and multimodal benchmarks.
Under this configuration, the total number of training tokens is approximately 200 billion, comprising 50 billion from language data and 150 billion from multimodal data.

\subsection{Post-Training}
\label{sec:post_training}

After the Native Multimodal Pre-Training, we apply a two-stage post-training strategy to further enhance the multimodal conversation and reasoning abilities of our models.
This strategy consists of Supervised Fine-Tuning (SFT) and Mixed Preference Optimization (MPO).
In the SFT phase, the model is trained to imitate the high-quality responses under positive supervision signals.
In the subsequent MPO phase, we introduce additional supervision from both positive and negative samples, thereby further improving its overall abilities.

\noindent\textbf{Supervised Fine-Tuning.}
In this phase, the techniques of random JPEG compression, square loss re-weighting, and multimodal data packing proposed in InternVL2.5~\cite{chen2024expanding} are also employed in the InternVL3 series.
The main advancement of the SFT phase in InternVL3 compared to InternVL2.5 lies in the use of higher-quality and more diverse training data.
Specifically, we further extend training samples for tool usage, 3D scene understanding, GUI operations, long context tasks, video understanding, scientific diagrams, creative writing, and multimodal reasoning.

\noindent\textbf{Mixed Preference Optimization.}
During Pre-training and SFT, the model is trained to predict the next token conditioned on previous ground-truth tokens.
However, during inference, the model predicts each token based on its own prior outputs. 
This discrepancy between ground-truth tokens and model-predicted tokens introduces a distribution shift, which can impair the model’s Chain-of-Thought (CoT) reasoning capabilities.
To mitigate this issue, we employ Mixed Preference Optimization (MPO)~\cite{wang2024mpo}, which introduces additional supervision from both positive and negative samples to align the model response distribution with the ground-truth distribution, thereby improving reasoning performance.
Specifically, the training objective of MPO is a combination of preference loss $\mathcal{L}_{p}$, quality loss $\mathcal{L}_{q}$, and generation loss $\mathcal{L}_{g}$, which can be formulated as follows:
\begin{equation}
    \mathcal{L}=
    w_{p} \mathcal{L}_{p}
    +
    w_{q} \mathcal{L}_{q}
    +
    w_{g} \mathcal{L}_{g}
    ,
    \label{eqn:final_loss}
\end{equation}
where $w_{*}$ represents the weight assigned to each loss component.
Specifically, the DPO loss~\cite{rafailov2024dpo} serves as the preference loss to enable the model to learn the relative preference between chosen and rejected responses:
\begin{equation}\small
    \mathcal{L}_p=
    -\log \sigma\left(\beta \log \frac{\pi_\theta\left(y_c \mid x\right)}{\pi_0\left(y_c \mid x\right)}-\beta \log \frac{\pi_\theta\left(y_r \mid x\right)}{\pi_0\left(y_r \mid x\right)}\right),
    \label{eqn:dpo_loss}
\end{equation}
where $\beta$ is the KL penalty coefficient,
and $x$, $y_c$, and $y_r$ are user query, chosen response, and rejected response, respectively.
The policy model $\pi_\theta$ is initialized from model $\pi_0$.
After that, the BCO loss~\cite{jung2024bco} is employed as the quality loss, which helps the model to understand the absolute quality of individual responses:
\begin{equation}
    \mathcal{L}_{q}=
    \mathcal{L}_{q}^+
    +
    \mathcal{L}_{q}^-,
\end{equation}
where $\mathcal{L}_{q}^+$ and $\mathcal{L}_{q}^-$ represent the loss for chosen and rejected responses, respectively.
They are calculated independently, requiring the model to differentiate the absolute quality of individual responses.
The loss terms are given by:
\begin{equation}\small
    \mathcal{L}_{q}^+=
    -\log \sigma\left(
        \beta \log \frac{\pi_\theta\left(y_c \mid x\right)}{\pi_0\left(y_c \mid x\right)}
        - \delta
    \right),
\end{equation}%
\begin{equation}\small
    \mathcal{L}_{q}^-=
    - 
    \log \sigma\left(
    -\left(
        \beta \log \frac{\pi_\theta\left(y_r \mid x\right)}{\pi_0\left(y_r \mid x\right)}
        - \delta
    \right) \right),
\end{equation}
where $\delta$ represents the reward shift, calculated as the moving average of previous rewards to stabilize training.
Finally, the LM loss is used as the generation loss to help the model learn the generation process of preferred responses.
The loss function is defined in Equation~\ref{eq:autoregressive-loss-text}.

\noindent\textbf{Data.}
For SFT data, we construct the training corpora based on those used in InternVL2.5~\cite{chen2024expanding} while introducing additional tool usage, 3D scene understanding, GUI operations, scientific diagrams, creative writing, and multimodal reasoning samples.
As a result, the number of training samples grows from 16.3M in InternVL2.5 to 21.7M in InternVL3.
For MPO data, we construct preference pairs based on the data pipeline and samples proposed in MMPR v1.2~\cite{wang2024mpo}, which cover a wide range of domains, including general visual question answering (VQA)~\cite{goyal2017vqav2,hudson2019gqa,marino2019okvqa,lu2021iconqa,wang2023allseeing,wang2024allseeingv2}, science~\cite{kembhavi2016ai2d,chen2024m3cot,lu2022scienceqa}, chart~\cite{masry2022chartqa,kafle2018dvqa,chang2022mapqa},  mathematics~\cite{lindstrom2022clevrmath,seo2015geos,cao2022geoqa_plus,lu2021geometry3k,kazemi2023geomverse,gao2023gllava,zhang2024mavis,shi2024mathv}, OCR~\cite{mathew2022infographicvqa,singh2019textvqa,biten2019stvqa,huang2019sroie,mishra2019ocrvqa}, and document~\cite{clark2017docqa}.
We use the SFT versions of InternVL3-8B, 38B, and 78B to generate rollouts. During the MPO phase, all models are trained on the same dataset, which comprises about 300K samples.

\subsection{Test-Time Scaling}

Test-Time Scaling has been shown to be an effective method to enhance the reasoning abilities of LLMs and MLLMs~\cite{snell2024test_time_scaling_efficient,mcaleese2024openai_critic,luo2024omegaprm,lightman2023prm800k,wang2023mathshepherd,fei2004learning,zhang2025qwen_prm,wang2025visualprm}.
In this work, we use the Best-of-N evaluation strategy and employ VisualPRM-8B~\cite{wang2025visualprm} as the critic model to select the best response for reasoning and mathematics evaluation.

\noindent\textbf{Visual Process Reward Model.}
VisualPRM first assigns a quality score to each step of the given solution and then averages these scores to obtain the overall score for this solution.
This process is formulated as a multi-turn chat task so that we can effectively leverage the generation ability of MLLMs.
The image $I$, question $q$, and the first step $s_0$ of the step-by-step solution $s=\{s_0,s_1,\cdots,s_n\} \in \mathcal{S}$ to this question are included in the first turn and a new step is presented in each subsequent turn.
During the training stage, the model is required to predict the correctness of the given step in each turn as follows:
\begin{equation}
    c_i \sim M(y_i \mid I,q, s_{\leq i}),
\end{equation}
where $c_i\in\{+,-\}$ denotes the correctness of $i$-th step.
During the inference stage, the score for each step is defined as the probability of generating ``$+$''.

\noindent\textbf{Data.}
VisualPRM400K~\cite{wang2025visualprm} is used to train VisualPRM, which is constructed based on multimodal questions collected from MMPR v1.2~\cite{wang2024mpo}. Following the data pipeline in VisualPRM400K, we further expand VisualPRM400K by sampling rollouts from the 8B and 38B variants of InternVL3.

\subsection{Infrastructure}

To facilitate model training, we extend the InternEVO framework~\cite{chen2024internevo}—originally designed to optimize the Zero Redundancy Optimizer (ZeRO) for large-scale LLM training—to support the training of our InternVL models. This extension enables efficient scaling to hundreds of billions of parameters across thousands of GPUs.
The enhanced framework introduces flexible and decoupled sharding strategies for the ViT, MLP, and LLM components, significantly improving training efficiency by overlapping communication and computation. It further supports a comprehensive range of parallelism strategies—including data, tensor, sequence, and pipeline parallelism—as well as their arbitrary combinations.

A key challenge in MLLM training is the imbalance in computational load caused by the varying proportions of visual and textual tokens. Such imbalances can lead to inefficiencies by overburdening either the ViT or LLM modules. To address this, we introduce a suite of techniques that dynamically balance computational workloads across modules, ensuring efficient and equitable resource utilization.

For InternVL models of varying scales, the extended InternEVO framework formulates an optimization objective that identifies the optimal configuration to minimize both memory consumption and communication overhead across different module dimensions. To support sequences of up to 32K tokens, our approach incorporates both head-parallel and sequence-parallel techniques, effectively overcoming scalability bottlenecks while preserving computational efficiency.
Compared to the training of InternVL2.5, the application of InternEVO in InternVL3 results in a training speedup of 50\% to 200\% for models of comparable size, given the same computational budget.

\section{Experiments}
In this section, we first compare the overall multimodal capabilities of InternVL3 with those of current advanced MLLMs using widely adopted multimodal benchmarks. Subsequently, we evaluate the performance of InternVL3 in various domains, including multimodal reasoning, mathematics, optical character recognition (OCR), chart and document understanding, multi-image understanding, real-world comprehension, comprehensive multimodal evaluation, multimodal hallucination evaluation, visual grounding, multimodal multilingual understanding, video understanding, and other multimodal tasks, most of which were tested using VLMEvalKit~\cite{duan2024vlmevalkit}.
Additionally, we provide a detailed evaluation of the language capabilities of InternVL3.
Finally, we analyze the advantages of several key modifications in InternVL3 compared to its predecessor, InternVL2.5, including the naive multimodal pre-training, the V2PE positional encoding, and the improvements brought by the post-training technique.

\subsection{Overall Comparison to Other Advanced MLLMs}

Figure~\ref{fig:overall_performance} provides a detailed assessment of InternVL3's performance across a diverse set of benchmarks, including MMMU~\cite{yue2023mmmu}, MathVista~\cite{lu2023mathvista}, AI2D~\cite{kembhavi2016ai2d}, ChartQA~\cite{masry2022chartqa}, DocVQA~\cite{mathew2021docvqa}, InfographicVQA~\cite{mathew2022infographicvqa}, HallusionBench~\cite{guan2023hallusionbench}, OCRBench~\cite{liu2023ocrbench}, and LongVideoBench~\cite{wu2024longvideobench}. Compared with previous models, InternVL3 demonstrates substantial improvements across a wide range of task categories. These advancements can be primarily attributed to enhanced training strategies, refined testing methodologies, and the expanded training corpus.

More specifically, InternVL3 achieves an impressive score of 72.2 on the MMMU benchmark, underscoring its superior capacity to manage complex multimodal challenges. Beyond its performance on MMMU, InternVL3 consistently outperforms earlier versions of the InternVL series on a variety of tasks, thereby emphasizing its broad applicability to real-world scenarios that require sophisticated multimodal comprehension and reasoning.

In addition to surpassing its open-source counterparts, InternVL3 exhibits competitive performance relative to leading closed-source commercial models, such as ChatGPT-4o-latest~\cite{chatgpt4o} and Claude-3.5 Sonnet~\cite{claude3series2024}. In many cases, the performance gap between InternVL3 and these proprietary models is notably narrowed—and in certain benchmarks, such as AI2D and ChartQA, InternVL3 even surpasses them. Nonetheless, our results further reveal that Gemini2.5 Pro~\cite{geminipro2.5} maintains a performance edge on select tasks (\eg, on HallusionBench), indicating that despite the notable progress in InternVL3, there remains room for further refinement of our InternVL series.

\subsection{Multimodal Reasoning and Mathematics}

\begin{table*}[t!]
\centering
\fontsize{8}{10}\selectfont 
\setlength\tabcolsep{4pt}
\renewcommand{\arraystretch}{0.9}
\begin{tabular}{l|ccccccc|c}
{Model}                           & {MMMU} & {MathVista} & {MathVision} & {MathVerse} & {DynaMath} & {WeMath} & {LogicVista} & {Overall} \\ 
\hline
LLaVA-OV-0.5B~\cite{li2024llavaov}       & 31.4          & 34.8               & $-$                   & $-$                  & $-$                 & $-$               & $-$                   & $-$                \\
InternVL2.5-1B~\cite{chen2024expanding}  & 41.2          & 47.1               & 21.1                & 16.4               & 5.6               & 11.1            & 26.0                & 24.1             \\
\rowcolor{gray!15}
InternVL3-1B                             & 43.4          & 45.8               & 18.8                & 18.7               & 5.8               & 13.4            & 29.8                & 25.1             \\
\rowcolor{gray!15}
\emph{\textcolor{blue}{w/ VisualPRM-Bo8}}~\cite{wang2025visualprm}                              & 55.4          & 62.1               & 21.7                & 28.9               & 13.4              & 28.5            & 34.9                & 35.0             \\
\hline
Aquila-VL-2B~\cite{gu2024aquilavl}       & 46.9          & 59.1               & 17.9                & 17.4               & 5.0               & 15.9            & 30.6                & 27.5             \\
Qwen2.5-VL-3B~\cite{bai2025qwen2_5}      & 51.2          & 61.2               & 21.9                & 31.2               & 13.2              & 22.9            & 40.3                & 34.6             \\
Ovis-2B~\cite{lu2024ovis}                & 45.6          & 64.1               & 17.7                & 29.4               & 10.0              & 9.9             & 34.7                & 30.2             \\
Ovis-4B~\cite{lu2024ovis}                & 49.0          & 69.6               & 21.5                & 38.5               & 18.0              & 16.9            & 35.3                & 35.5             \\
InternVL2.5-2B~\cite{chen2024expanding}  & 43.2          & 51.1               & 14.0                & 22.3               & 4.4               & 8.0             & 27.3                & 24.3             \\
InternVL2.5-4B~\cite{chen2024expanding}  & 51.8          & 64.1               & 18.4                & 27.7               & 15.2              & 21.2            & 34.2                & 33.2             \\
\rowcolor{gray!15}
InternVL3-2B                             & 48.6          & 57.0               & 21.7                & 25.3               & 14.6              & 22.4            & 36.9                & 32.4             \\
\rowcolor{gray!15}
\emph{\textcolor{blue}{w/ VisualPRM-Bo8}}~\cite{wang2025visualprm}                              & 57.8          & 70.5               & 26.6                & 36.7               & 21.4              & 38.5            & 40.5                & 41.7             \\
\hline
LLaVA-OV-7B~\cite{li2024llavaov}         & 47.9          & 58.6               & 18.3                & 19.3               & 9.0               & 20.9            & 33.3                & 29.6             \\
MiniCPM-V2.6~\cite{yao2024minicpm}       & 49.8          & 60.8               & 23.4                & 18.9               & 9.8               & 16.4            & 27.5                & 29.5             \\
MiniCPM-o2.6~\cite{yao2024minicpm}       & 50.9          & 73.3               & 21.7                & 35.0               & 10.4              & 25.2            & 36.0                & 36.1             \\
Ovis-8B~\cite{lu2024ovis}                & 57.4          & 71.8               & 25.9                & 42.3               & 20.4              & 27.2            & 39.4                & 40.6             \\
Qwen2.5-VL-8B~\cite{bai2025qwen2_5}      & 55.0          & 67.8               & 25.4                & 41.1               & 21.0              & 35.2            & 44.1                & 41.4             \\
InternVL2.5-8B~\cite{chen2024expanding}  & 56.2          & 64.5               & 17.0                & 22.8               & 9.4               & 23.5            & 36.0                & 32.8             \\
\rowcolor{gray!15}
InternVL3-8B                             & 62.7          & 71.6               & 29.3                & 39.8               & 25.5              & 37.1            & 44.1                & 44.3             \\
\rowcolor{gray!15}
\emph{\textcolor{blue}{w/ VisualPRM-Bo8}}~\cite{wang2025visualprm}                              & 66.0          & 75.2               & 37.5                & 46.3               & 28.5              & 48.1            & 49.7                & 50.2             \\
\hline
\rowcolor{gray!15}
InternVL3-9B                             & 57.7          & 71.5               & 27.6                & 35.3               & 26.7              & 33.8            & 49.2                & 43.1             \\
\rowcolor{gray!15}
\emph{\textcolor{blue}{w/ VisualPRM-Bo8}}~\cite{wang2025visualprm}                              & 63.7          & 76.2               & 33.9                & 45.8               & 29.1              & 46.6            & 50.6                & 49.4             \\
\hline
Ovis2-16B~\cite{lu2024ovis}              & 60.7          & 73.7               & 30.1                & 45.8               & 26.3              & 45.0            & 47.4                & 47.0             \\
InternVL2.5-26B~\cite{chen2024expanding} & 60.7          & 68.2               & 23.4                & 24.0               & 11.4              & 30.9            & 39.6                & 36.9             \\
\rowcolor{gray!15}
InternVL3-14B                            & 67.1          & 75.1               & 37.2                & 44.4               & 31.3              & 43.0            & 51.2                & 49.9             \\
\rowcolor{gray!15}
\emph{\textcolor{blue}{w/ VisualPRM-Bo8}}~\cite{wang2025visualprm}                             & 69.3          & 77.9               & 40.1                & 47.7               & 33.1              & 52.0            & 56.2                & 53.8             \\
\hline
Cambrian-34B~\cite{tong2024cambrian}     & 49.7          & 53.2               & $-$                   & $-$                  & $-$                 & $-$               & $-$                   & $-$                \\
VILA-1.5-40B~\cite{lin2024vila}          & 55.1          & 49.5               & $-$                   & $-$                  & $-$                 & $-$               & $-$                   & $-$                \\
Ovis2-34B~\cite{lu2024ovis}              & 66.7          & 76.1               & 31.9                & 50.1               & 27.5              & 51.9            & 49.9                & 50.6             \\
InternVL2.5-38B~\cite{chen2024expanding} & 63.9          & 71.9               & 32.2                & 36.9               & 20.0              & 38.3            & 47.9                & 44.4             \\
\rowcolor{gray!15}
InternVL3-38B                            & 70.1          & 75.1               & 34.2                & 48.2               & 35.3              & 48.6            & 58.4                & 52.8             \\
\rowcolor{gray!15}
\emph{\textcolor{blue}{w/ VisualPRM-Bo8}}~\cite{wang2025visualprm}                              & 71.0          & 79.4               & 41.8                & 54.2               & 36.1              & 55.2            & 58.4                & 56.6             \\
\hline
GPT-4o-20241120~\cite{gpt4v}             & 70.7          & 60.0               & 31.2                & 40.6               & 34.5              & 45.8            & 52.8                & 47.9             \\
Claude-3.7-Sonnet~\cite{claude3series2024}& 75.0          & 66.8               & 41.9                & 46.7               & 39.7              & 49.3            & 58.2                & 53.9             \\
Gemini-2.0-Flash~\cite{gemini2_0}        & 72.6          & 70.4               & 43.6                & 47.8               & 42.1              & 47.4            & 52.3                & 53.7             \\
Gemini-2.0-Pro~\cite{gemini2_0pro}       & 69.9          & 71.3               & 48.1                & 67.3               & 43.3              & 56.5            & 53.2                & 58.5             \\
LLaVA-OV-72B~\cite{li2024llavaov}        & 55.7          & 67.1               & 25.3                & 27.2               & 15.6              & 32.0            & 40.9                & 37.7             \\
QvQ-72B-Preview~\cite{qvq-72b-preview}   & 70.3          & 70.3               & 34.9                & 48.2               & 30.7              & 39.0            & 58.2                & 50.2             \\
Qwen2.5-VL-72B~\cite{bai2025qwen2_5}     & 68.2          & 74.2               & 39.3                & 47.3               & 35.9              & 49.1            & 55.7                & 52.8             \\
InternVL2.5-78B~\cite{chen2024expanding} & 70.0          & 72.3               & 32.2                & 39.2               & 19.2              & 39.8            & 49.0                & 46.0             \\
\rowcolor{gray!15}
InternVL3-78B                            & 72.2          & 79.0               & 43.1                & 51.0               & 35.1              & 46.1            & 55.9                & 54.6             \\
\rowcolor{gray!15}
\emph{\textcolor{blue}{w/ VisualPRM-Bo8}}~\cite{wang2025visualprm}                              & 72.2          & 80.5               & 40.8                & 54.2               & 37.3              & 52.4            & 57.9                & 56.5             \\ 
\end{tabular}

\caption{\textbf{Comparison of multimodal reasoning and mathematical performance.}
MMMU~\cite{yue2023mmmu} is a multidisciplinary reasoning benchmark.
MathVista~\cite{lu2023mathvista}, MathVision~\cite{wang2024mathvision}, MathVerse~\cite{zhang2024mathverse}, DynaMath~\cite{zou2024dynamath}, and WeMath~\cite{qiao2024wemath} are mathematics benchmarks. For MathVerse, we report the performance on Vision-Only split.
LogicVista~\cite{xiao2024logicvista} is a logical reasoning benchmark.
Part of the results are collected from the OpenCompass leaderboard~\cite{opencompass2023}.
The overall score is the average score of the above benchmarks.
``w/ VisualPRM-Bo8'' denotes that the model is evaluated with Best-of-8 settings, where VisualPRM~\cite{wang2025visualprm} serves as the critic model.
}
\label{tab:benchmark_math_reasoning}
\end{table*}

\label{sec:math_reasoning_datasets}
To comprehensively evaluate the multimodal reasoning and mathematical capabilities of InternVL3, we conduct experiments on a series of benchmarks, including MMMU~\cite{yue2023mmmu} for multidisciplinary reasoning, MathVista~\cite{lu2023mathvista}, MathVision~\cite{wang2024mathvision}, MathVerse~\cite{zhang2024mathverse} for mathematical reasoning, as well as DynaMath~\cite{zou2024dynamath}, WeMath~\cite{qiao2024wemath} and LogicVista~\cite{xiao2024logicvista} for complementary evaluation on logical reasoning.

As shown in Table~\ref{tab:benchmark_math_reasoning}, InternVL3 exhibits strong performance across all tested benchmarks. Specifically, on the MMMU benchmark, InternVL3-based models consistently outperform smaller-scale competitors. For instance, with increasing model size, InternVL3‑78B reaches a score over 72 on MMMU, indicating robust understanding and reasoning capability in handling abstract multidisciplinary concepts. In the mathematical domain, InternVL3 demonstrates significant gains across various benchmarks. On MathVista, InternVL3‑78B records a performance close to 79.0, while on MathVision and MathVerse, the results are also competitive, evidencing the model’s enhanced ability to tackle challenging mathematical problems. Furthermore, performance on  DynaMath, WeMath, and LogicVista consistently improves with scaling. The overall score—a mean calculated across all benchmarks—shows that InternVL3 models achieve a balanced enhancement across different aspects, surpassing many of the preceding open-source methods.

A notable characteristic of InternVL3 is the efficiency of the best-of-N evaluation strategy~\cite{wang2025visualprm}. When applying this method, even models with relatively smaller parameter sizes (\eg, InternVL3‑1B and InternVL3‑2B) exhibit substantial improvements in reasoning performance. Specifically, in the Vision-Only split of MathVerse, the best-of‑8 strategy leads to increases of approximately 6.0 and 3.2 percentage points for InternVL3‑38B and InternVL3‑78B, respectively. This improvement underscores the effectiveness of test-time scaling.

\subsection{OCR, Chart, and Document Understanding}

\label{sec:ocr_datasets}

\begin{table*}[t!]
\centering
{\fontsize{8}{10}\selectfont 
\renewcommand{\arraystretch}{0.95}
\setlength\tabcolsep{1.5pt}
\newcommand{\TextVQA}{\makecell{TextVQA\\(val)}}
\newcommand{\ChartQA}{\makecell{ChartQA\\(test avg)}}
\newcommand{\DocVQA}{\makecell{DocVQA\\(test)}}
\newcommand{\InfoVQA}{\makecell{InfoVQA\\(test)}}
\newcommand{\TableVQA}{\makecell{TableVQA\\Bench}}
\newcommand{\CharXiv}{\makecell{CharXiv\\(RQ / DQ)}}
\newcommand{\SEEDTP}{\makecell{SEED-2\\Plus}}
\newcommand{\VCREN}{\makecell{VCR-EN-Easy\\(EM / Jaccard)}}
\newcommand{\VCRZH}{\makecell{VCR-ZH\\(EM/Jaccard)}}
\newcommand{\OCRBench}{\makecell{OCR\\Bench}}
\newcommand{\AITD}{\makecell{AI2D\\(w / wo M)}}
\newcommand{\lsp}{--\ \ \ \ \ \ }
\newcommand{\rsp}{\ \ \ \ \ --~}
\newcommand{\TODO}{\textcolor{red}{TODO}}
\newcommand{\newicon}{\includegraphics[height=1em]{figure/new.pdf}}
\begin{tabular}{l|cccccccccc}
Model Name                                 & \AITD       & \ChartQA & \TextVQA & \DocVQA & \InfoVQA & \OCRBench & \SEEDTP & \CharXiv    & \VCREN       & Overall \\
\hline
LLaVA-OneVision-0.5B~\cite{li2024llavaov}  & 57.1 / \lsp & 61.4     & --       & 70.0    & 41.8     & 565       & --      & --          & --           & --      \\
InternVL2-1B~\cite{chen2024far}            & 64.1 / 70.5 & 72.9     & 70.5     & 81.7    & 50.9     & 754       & 54.3    & 18.1 / 30.7 & 21.5 / 48.4  & 54.9    \\
InternVL2.5-1B~\cite{chen2024expanding}    & 69.3 / 77.8 & 75.9     & 72.0     & 84.8    & 56.0     & 785       & 59.0    & 19.0 / 38.4 & 91.5 / 97.0  & 68.3    \\
\rowcolor{gray!15}
InternVL3-1B                               & 69.4 / 78.3 & 75.3     & 74.1     & 81.9    & 53.7     & 790       & 58.2    & 21.0 / 47.1 & 89.3 / 96.2  & 68.6    \\
Qwen2-VL-2B~\cite{wang2024qwen2vl}         & 74.7 / 84.6 & 73.5     & 79.7     & 90.1    & 65.5     & 809       & 62.4    & --          & 81.5 / \lsp  & --      \\
Qwen2.5-VL-3B~\cite{bai2025qwen2_5}        & 81.6 / \lsp & 84.0     & 79.3     & 93.9    & 77.1     & 797       & 67.6    & 31.3 / 58.6 & --           & --      \\
Aquila-VL-2B~\cite{gu2024aquilavl}         & 75.0 / \lsp & 76.5     & 76.4     & 85.0    & 58.3     & 772       & 63.0    & --          & 70.0 / \lsp  & --      \\
InternVL2-2B~\cite{chen2024far}            & 74.1 / 82.3 & 76.2     & 73.4     & 86.9    & 58.9     & 784       & 60.0    & 21.0 / 40.6 & 32.9 / 59.2  & 62.0    \\
InternVL2.5-2B~\cite{chen2024expanding}    & 74.9 / 83.5 & 79.2     & 74.3     & 88.7    & 60.9     & 804       & 60.9    & 21.3 / 49.7 & 93.2 / 97.6  & 72.1    \\
\rowcolor{gray!15}
InternVL3-2B                               & 78.7 / 87.4 & 80.2     & 77.0     & 88.3    & 66.1     & 835       & 64.6    & 28.3 / 54.7 & 91.2 / 96.9  & 74.7    \\
\hline
Ovis1.6-Gemma2-9B~\cite{lu2024ovis}        & 84.4 / \lsp & --       & --       & --      & --       & 830       & --      & --          & --           & --      \\
MiniCPM-V2.6~\cite{yao2024minicpm}         & 82.1 / \lsp & 82.4     & 80.1     & 90.8    & --       & 852       & 65.7    & 31.0 / 57.1 & 73.9 / 85.7  & --      \\
Molmo-7B-D~\cite{deitke2024molmo}          & \rsp / 93.2 & 84.1     & 81.7     & 92.2    & 72.6     & 694       & --      & --          & --           & --      \\
Qwen2-VL-7B~\cite{wang2024qwen2vl}         & 83.0 / 92.1 & 83.0     & 84.3     & 94.5    & 76.5     & 866       & 69.0    & --          & 89.7 / 93.8  & --      \\
Qwen2.5-VL-7B~\cite{bai2025qwen2_5}        & 83.9 / \lsp & 87.3     & 84.9     & 95.7    & 82.6     & 864       & 70.4    & 42.5/73.9   & --           & --      \\
InternVL2-8B~\cite{chen2024far}            & 83.8 / 91.7 & 83.3     & 77.4     & 91.6    & 74.8     & 794       & 67.5    & 31.2 / 56.1 & 37.9 / 61.5  & 69.7    \\
InternVL2.5-8B~\cite{chen2024expanding}    & 84.5 / 92.8 & 84.8     & 79.1     & 93.0    & 77.6     & 822       & 69.7    & 32.9 / 68.6 & 92.6 / 97.4  & 79.6    \\
\rowcolor{gray!15}
InternVL3-8B                               & 85.2 / 92.6 & 86.6     & 80.2     & 92.7    & 76.8     & 880       & 69.7    & 37.6 / 73.6 & 94.5 / 98.1  & 81.3    \\
\rowcolor{gray!15}
InternVL3-9B                               & 84.6 / 92.9 & 86.2     & 79.4     & 93.6    & 79.6     & 877       & 68.8    & 38.0 / 72.5 & 94.2 / 97.9  & 81.3    \\
\rowcolor{gray!15}
InternVL3-14B                              & 86.0 / 93.7 & 87.3     & 80.5     & 94.1    & 83.6     & 875       & 70.3    & 43.1 / 82.2 & 94.8 / 98.2  & 83.4    \\
\hline
InternVL-Chat-V1.5~\cite{chen2024far}      & 80.7 / 89.8 & 83.8     & 80.6     & 90.9    & 72.5     & 724       & 66.3    & 29.2 / 58.5 & 14.7 / 51.4  & 65.9    \\
InternVL2-26B~\cite{chen2024far}           & 84.5 / 92.5 & 84.9     & 82.3     & 92.9    & 75.9     & 825       & 67.6    & 33.4 / 62.4 & 74.5 / 86.7  & 76.7    \\
InternVL2.5-26B~\cite{chen2024expanding}   & 86.4 / 94.4 & 87.2     & 82.4     & 94.0    & 79.8     & 852       & 70.8    & 35.9 / 73.5 & 94.4 / 98.0  & 81.8    \\
Qwen2.5-VL-32B~\cite{bai2025qwen2_5}       & --          & --       & --       & 94.8    & 83.4     & --        & --      & --          & --           & --      \\
Cambrian-34B~\cite{tong2024cambrian}       & 79.5 / \lsp & 75.6     & 76.7     & 75.5    & 46.0     & 600       & --      & 27.3 / 59.7 & 79.7 / 89.3  & --      \\
VILA-1.5-40B~\cite{lin2024vila}            & 69.9 / \lsp & 67.2     & 73.6     & --      & --       & 460       & --      & 24.0 / 38.7 & --           & --      \\
InternVL2-40B~\cite{chen2024far}           & 86.6 / 94.5 & 86.2     & 83.0     & 93.9    & 78.7     & 837       & 69.2    & 32.3 / 66.0 & 84.7 / 92.6  & 79.3    \\
InternVL2.5-38B~\cite{chen2024expanding}   & 87.6 / 95.1 & 88.2     & 82.7     & 95.3    & 83.6     & 842       & 71.2    & 42.4 / 79.6 & 94.7 / 98.2  & 83.6    \\
\rowcolor{gray!15}
InternVL3-38B                              & 88.9 / 95.5 & 89.2     & 83.9     & 95.4    & 85.0     & 886       & 71.6    & 46.4 / 87.2 & 96.1 / 98.7  & 85.5    \\
\hline
GPT-4V~\cite{gpt4v}                        & 78.2 / 89.4 & 78.5     & 78.0     & 88.4    & 75.1     & 645       & 53.8    & 37.1 / 79.9 & 52.0 / 65.4  & 70.0    \\
GPT-4o-20240513~\cite{gpt4v}               & 84.6 / 94.2 & 85.7     & 77.4     & 92.8    & 79.2     & 736       & 72.0    & 47.1 / 84.5 & 91.6 / 96.4  & 81.6    \\
Claude-3-Opus~\cite{claude3series2024}     & 70.6 / 88.1 & 80.8     & 67.5     & 89.3    & 55.6     & 694       & 44.2    & 30.2 / 71.6 & 62.0 / 77.7  & 67.3    \\
Claude-3.5-Sonnet~\cite{claude3series2024} & 81.2 / 94.7 & 90.8     & 74.1     & 95.2    & 74.3     & 788       & 71.7    & 60.2 / 84.3 & 63.9 / 74.7  & 78.7    \\
Gemini-1.5-Pro~\cite{reid2024gemini1_5}    & 79.1 / 94.4 & 87.2     & 78.8     & 93.1    & 81.0     & 754       & --      & 43.3 / 72.0 & 62.7 / 77.7  & --      \\
LLaVA-OneVision-72B~\cite{li2024llavaov}   & 85.6 / \lsp & 83.7     & 80.5     & 91.3    & 74.9     & 741       & --      & --          & --           & --      \\
NVLM-D-72B~\cite{dai2024nvlm}              & 85.2 / 94.2 & 86.0     & 82.1     & 92.6    & --       & 853       & --      & --          & --           & --      \\
Molmo-72B~\cite{deitke2024molmo}           & \rsp / 96.3 & 87.3     & 83.1     & 93.5    & 81.9     & --        & --      & --          & --           & --      \\
Qwen2-VL-72B~\cite{wang2024qwen2vl}        & 88.1 / \lsp & 88.3     & 85.5     & 96.5    & 84.5     & 877       & --      & --          & 91.3 / 94.6  & --      \\
Qwen2.5-VL-72B~\cite{bai2025qwen2_5}       & 88.7 / \lsp & 89.5     & 83.5     & 96.4    & 87.3     & 885       & 73.0    & 49.7 / 87.4 & --           & --      \\
InternVL2-Llama3-76B~\cite{chen2024far}    & 87.6 / 94.8 & 88.4     & 84.4     & 94.1    & 82.0     & 839       & 69.7    & 38.9 / 75.2 & 83.2 / 91.3  & 81.1    \\
InternVL2.5-78B~\cite{chen2024expanding}   & 89.1 / 95.7 & 88.3     & 83.4     & 95.1    & 84.1     & 854       & 71.3    & 42.4 / 82.3 & 95.7 / 94.5  & 83.9    \\
\rowcolor{gray!15}
InternVL3-78B                              & 89.7 / 96.0 & 89.7     & 84.3     & 95.4    & 86.5     & 906       & 71.9    & 46.0 / 85.1 & 96.0 / 98.6  & 85.8    \\
\end{tabular}
}
\caption{\textbf{Comparison of OCR, chart, and document understanding performance.} We evaluate OCR-related capabilities across 9 benchmarks, including AI2D~\cite{kembhavi2016ai2d}, ChartQA~\cite{masry2022chartqa}, TextVQA~\cite{singh2019textvqa}, DocVQA~\cite{mathew2021docvqa}, InfoVQA~\cite{mathew2022infographicvqa}, OCRBench~\cite{liu2023ocrbench}, SEED-2-Plus~\cite{li2024seedbench2plus}, CharXiv~\cite{wang2024charxiv}, and VCR~\cite{zhang2024vcr}.
Part of results are collected from \cite{dubey2024llama3, deitke2024molmo, claude3series2024, wang2024charxiv, zhang2024vcr} and the OpenCompass leaderboard~\cite{opencompass2023}.
}
\vspace{-0.5em}
\label{tab:benchmark_ocr}
\end{table*}

To assess the model’s integrated vision–language understanding in tasks involving text, document, and chart comprehension, we perform a comprehensive evaluation over nine benchmarks, including AI2D~\cite{kembhavi2016ai2d}, ChartQA~\cite{masry2022chartqa}, TextVQA~\cite{singh2019textvqa}, DocVQA~\cite{mathew2021docvqa}, InfoVQA~\cite{mathew2022infographicvqa}, OCRBench~\cite{liu2023ocrbench}, SEED-2-Plus~\cite{li2024seedbench2plus}, CharXiv~\cite{wang2024charxiv}, and VCR~\cite{zhang2024vcr}. As illustrated in Table~\ref{tab:benchmark_ocr}, the InternVL3 series not only maintains robust performance across these benchmarks but also demonstrates competitive or superior results when compared to other open-source and closed-source counterparts.

At the 1B scale, InternVL3-1B achieves performance that is roughly on par with previous lower-scale models. 
At the 2B scale, InternVL3-2B not only improves its absolute scores—for instance, reaching 78.7/87.4 on AI2D and 88.3 on DocVQA—but also exhibits a performance edge over similarly parameterized models such as Qwen2-VL-2B~\cite{wang2024qwen2vl}. Although its TextVQA performance (77.0) remains comparable to that of Qwen2-VL-2B, the enhancements in document and chart understanding suggest that the proposed native multimodal pre-training are particularly effective in tasks requiring precise visual–textual integration.

The benefits of the new pre-training protocol become even more pronounced at larger scales. Mid-scale models like InternVL3-8B and InternVL3-9B deliver substantial gains, with InternVL3-8B achieving 85.2/92.6 on AI2D, 92.7 on DocVQA, and VCR scores of 94.5/98.1. Moreover, when compared with heavyweight systems such as Qwen2-VL-72B~\cite{wang2024qwen2vl} or even closed-source models like GPT-4o-20240513~\cite{gpt4v}, the high-scale variants of InternVL3—particularly InternVL3-38B and InternVL3-78B—push the envelope further. For instance, InternVL3-78B attains a remarkable OCRBench score of 906 and VCR scores of 96.0/98.6, clearly surpassing the corresponding metrics of comparable models.

\subsection{Multi-Image Understanding}

\label{sec:multi_image_datasets}

\begin{table*}[t!]
\scriptsize
\centering
{\fontsize{8}{10}\selectfont 
\renewcommand{\arraystretch}{0.95}
\setlength\tabcolsep{1.3pt}
\newcommand{\MMIU}{\makecell{MMIU}}
\newcommand{\Muir}{\makecell{Muir\\Bench}} 
\newcommand{\BLINK}{\makecell{BLINK\\(val)}}
\newcommand{\Mantis}{\makecell{Mantis\\Eval}} 
\newcommand{\RWQA}{\makecell{RealWorld\\QA}} 
\newcommand{\MMERW}{\makecell{MME-RW\\(EN)}} 
\newcommand{\RBench}{\makecell{R-Bench\\(test)}}
\newcommand{\TaskMe}{\makecell{TaskMe-\\Anything}}
\newcommand{\MMT}{\makecell{MMT\\(val)}}
\newcommand{\WILDV}{\makecell{WildVision\\(win rate)}}
\newcommand{\MIRB}{\makecell{MIRB\\(avg)}}
\newcommand{\RB}{\makecell{R-Bench\\(dis)}}
\newcommand{\TODO}{\textcolor{red}{TODO}}
\newcommand{\lsp}{--\ \ \ \ \ \ }
\newcommand{\rsp}{\ \ \ \ \ --~}
\begin{tabular}{l|ccccccc|ccccc}
Model Name                                 & \BLINK & \Mantis & \MMIU & \Muir & \MMT & \MIRB & Overall &\RWQA  & \MMERW & \WILDV  & \RB  & Overall \\
\hline
LLaVA-OneVision-0.5B~\cite{li2024llavaov}  & 52.1   & 39.6    & --    & 25.5  & --   & --    & --      & 55.6  & --     & --      & --   & --      \\
InternVL2-1B~\cite{chen2024far}            & 38.6   & 46.1    & 37.3  & 29.3  & 49.5 & 31.5  & 38.7    & 50.3  & 40.2   & 17.8    & 55.6 & 41.0    \\
InternVL2.5-1B~\cite{chen2024expanding}    & 42.0   & 51.2    & 38.5  & 29.9  & 50.3 & 35.6  & 41.3    & 57.5  & 44.2   & 43.4    & 59.0 & 51.0    \\
\rowcolor{gray!15}
InternVL3-1B                               & 42.9   & 50.2    & 39.3  & 31.2  & 52.9 & 36.1  & 42.1    & 58.2  & 46.0   & 43.8    & 60.4 & 52.1    \\
Qwen2-VL-2B~\cite{wang2024qwen2vl}         & 44.4   & --      & --    & --    & 55.1 & --    & --      & 62.6  & --     & --      & --   & --      \\
Qwen2.5-VL-3B~\cite{bai2025qwen2}          & 47.6   & --      & --    & 47.7    & -- & --    & --      & 65.4  & 53.1   & --      & --   & --      \\
InternVL2-2B~\cite{chen2024far}            & 43.8   & 48.4    & 39.8  & 32.5  & 50.4 & 32.1  & 41.2    & 57.3  & 47.3   & 31.8    & 56.8 & 48.3    \\
InternVL2.5-2B~\cite{chen2024expanding}    & 44.0   & 54.8    & 43.5  & 40.6  & 54.5 & 36.4  & 45.6    & 60.1  & 48.8   & 44.2    & 62.2 & 53.8    \\
\rowcolor{gray!15}
InternVL3-2B                               & 50.3   & 65.9    & 43.0  & 38.8  & 59.5 & 42.9  & 50.1    & 64.3  & 53.8   & 48.8    & 67.5 & 58.6    \\
\hline
Qwen2-VL-7B~\cite{wang2024qwen2vl}         & 53.2   & --      & --    & --    & 64.0 & --    & --      & 70.1  & 56.5   & --      & 64.0 & --      \\
Qwen2.5-VL-7B~\cite{bai2025qwen2}          & 56.4   & --      & --    & 59.6  & --   & --    & --      & 68.5  & 57.4   & --      & --   & --      \\
MiniCPM-V2.6~\cite{yao2024minicpm}         & 53.0   & 69.0    & --    & --    & 60.8 & --    & --      & 65.0  & --     & --      & --   & --      \\
InternVL2-8B~\cite{chen2024far}            & 50.9   & 65.4    & 42.0  & 48.7  & 60.0 & 50.0  & 52.8    & 64.4  & 53.5   & 54.4    & 67.9 & 60.1    \\
InternVL2.5-8B~\cite{chen2024expanding}    & 54.8   & 67.7    & 46.7  & 51.1  & 62.3 & 52.5  & 55.9    & 70.1  & 59.1   & 62.0    & 70.1 & 65.3    \\
\rowcolor{gray!15}
InternVL3-8B                               & 55.5   & 70.1    & 46.8  & 55.0  & 65.0 & 56.8  & 58.2    & 70.8  & 62.0   & 69.8    & 74.1 & 69.2    \\
\rowcolor{gray!15}
InternVL3-9B                               & 58.6   & 70.1    & 50.4  & 51.4  & 65.4 & 58.6  & 59.1    & 70.5  & 61.3   & 63.8    & 70.3 & 66.5    \\
\rowcolor{gray!15}
InternVL3-14B                              & 60.3   & 76.0    & 50.9  & 56.2  & 70.3 & 59.3  & 62.2    & 70.7  & 64.0   & 69.8    & 69.3 & 68.5    \\
\hline
InternVL-Chat-V1.5~\cite{chen2024far}      & 46.6   & 66.8    & 37.4  & 38.5  & 58.0 & 50.3  & 49.6    & 66.0  & 49.4   & 56.6    & 67.9 & 60.0    \\
InternVL2-26B~\cite{chen2024far}           & 56.2   & 69.6    & 42.6  & 50.6  & 60.6 & 53.7  & 55.6    & 68.3  & 58.7   & 62.2    & 70.1 & 64.8    \\
InternVL2.5-26B~\cite{chen2024expanding}   & 61.8   & 75.6    & 49.4  & 61.1  & 66.9 & 55.7  & 61.8    & 74.5  & 61.8   & 65.2    & 72.9 & 68.6    \\
Cambrian-34B~\cite{tong2024cambrian}       & --     & --      & --    & --    & --   & --    & --      & 67.8  & 44.1   & --      & --   & --      \\
InternVL2-40B~\cite{chen2024far}           & 57.2   & 71.4    & 47.9  & 54.4  & 66.2 & 55.2  & 58.7    & 71.8  & 61.8   & 63.2    & 73.3 & 67.5    \\
InternVL2.5-38B~\cite{chen2024expanding}   & 63.2   & 78.3    & 55.3  & 62.7  & 70.0 & 61.2  & 65.1    & 73.5  & 64.0   & 66.4    & 72.1 & 69.0    \\
\rowcolor{gray!15}
InternVL3-38B                              & 64.0   & 77.9    & 57.4  & 63.8  & 71.8 & 62.3  & 66.2    & 75.6  & 67.3   & 71.6    & 73.3 & 72.0    \\
\hline
GPT-4V~\cite{gpt4v}                        & 54.6   & 62.7    & --    & 62.3  & 64.3 & 53.1  & --      & 61.4  & --     & 71.8    & 65.6 & --      \\
GPT-4o-20240513~\cite{gpt4v}               & 68.0   & --      & 55.7  & 68.0  & 65.4 & --    & --      & 75.4  & 45.2   & 80.6    & 77.7 & 69.7    \\
Claude-3.5-Sonnet~\cite{claude3series2024} & --     & --      & 53.4  & --    & --   & --    & --      & 60.1  & 51.6   & --      & --   & --      \\
Gemini-1.5-Pro~\cite{reid2024gemini1_5}    & --     & --      & 53.4  & --    & 64.5 & --    & --      & 67.5  & 38.2   & --      & --   & --      \\
LLaVA-OneVision-72B~\cite{li2024llavaov}   & 55.4   & 77.6    & --    & 54.8  & --   & --    & --      & 71.9  & --     & --      & --   & --      \\
Qwen2-VL-72B~\cite{wang2024qwen2vl}        & --     & --      & --    & --    & 71.8 & --    & --      & 77.8  & --     & --      & --   & --      \\
Qwen2.5-VL-72B~\cite{bai2025qwen2}         & 64.4   & --      & --    & 70.7  & --   & --    & --      & 75.7  & 63.2   & --      & --   & --      \\
InternVL2-Llama3-76B~\cite{chen2024far}    & 56.8   & 73.7    & 44.2  & 51.2  & 67.4 & 58.2  & 58.6    & 72.2  & 63.0   & 65.8    & 74.1 & 68.8    \\
InternVL2.5-78B~\cite{chen2024expanding}   & 63.8   & 77.0    & 55.8  & 63.5  & 70.8 & 61.1  & 65.3    & 78.7  & 62.9   & 71.4    & 77.2 & 72.6    \\
\rowcolor{gray!15}
InternVL3-78B                              & 66.3   & 79.3    & 60.4  & 64.5  & 73.2 & 64.3  & 68.0    & 78.0  & 65.4   & 73.6    & 77.4 & 73.6    \\
\end{tabular}
}
\caption{\textbf{Comparison of multi-image and real-world understanding performance. }
Multi-image benchmarks include BLINK~\cite{fu2024blink}, Mantis-Eval~\cite{jiang2024mantis}, MMIU~\cite{meng2024mmiu}, MuirBench~\cite{wang2024muirbench}, MMT-Bench~\cite{mmtbench}, and MIRB~\cite{zhao2024mirb}.
Real-world benchmarks encompass RealWorldQA~\cite{realworldqa}, MME-RealWorld~\cite{zhang2024mme}, WildVision~\cite{lu2024wildvision}, and R-Bench~\cite{li2024r}.
Part of the results are sourced from the benchmark papers and the OpenCompass leaderboard~\cite{opencompass2023}.
}
\label{tab:benchmark_multi_image_real_world}
\end{table*}

we evaluate the multi-image relation perception and understanding capabilities of InternVL3 across a suite of widely recognized benchmarks, including BLINK~\cite{fu2024blink}, Mantis-Eval~\cite{jiang2024mantis}, MMIU~\cite{meng2024mmiu}, MuirBench~\cite{wang2024muirbench}, MMT-Bench~\cite{mmtbench}, and MIRB~\cite{zhao2024mirb}, as presented in Table~\ref{tab:benchmark_multi_image_real_world}. These benchmarks comprehensively assess skills such as cross-image reasoning and context integration, all of which are crucial for effective multimodal interaction.

InternVL3 consistently outperforms its earlier counterparts across different parameter scales. For instance, at the 1B scale, InternVL3-1B exhibits a modest yet consistent improvement over preceding models, achieving a BLINK score of 42.9 and an MMT-Bench score of 52.9. The performance gains become even more pronounced at the 2B scale; InternVL3-2B attains a remarkable 65.9 on Mantis-Eval, representing an improvement of over 11 points relative to InternVL2.5-2B, and also boosts its MMT-Bench performance to 59.5. Such enhancements indicate that the advanced pre-training strategies and enhanced training datasets in InternVL3 significantly elevate its capability to capture and reason over inter-image relationships.

At higher scales, the trend continues. InternVL3-8B and its subsequent larger variants not only secure steady improvements on BLINK and MMT-Bench but also demonstrate substantial gains on the MIRB and MuirBench benchmarks. In particular, InternVL3-78B reaches a BLINK score of 66.3 and an MMT-Bench score of 73.2, positioning it as a competitive alternative to leading closed-source models like GPT-4o. These results suggest that the learning multimodal capabilities via native multimodal pre-training  and the scaling of model parameters are key contributors to the elevated performance observed across diverse evaluation settings.
Despite these encouraging outcomes, a noticeable performance gap between our InternVL3 and other MLLMs like Qwen2.5-VL still exists on certain benchmarks, such as MuirBench, implying that future work may benefit from further enhancements in training data curation and additional model refinements.

\subsection{Real-World Comprehension}

\label{sec:real_world_datasets}

We evaluate the InternVL3 series on four real-world comprehension benchmarks—RealWorldQA~\cite{realworldqa}, MME-RealWorld~\cite{zhang2024mme}, WildVision~\cite{lu2024wildvision}, and R-Bench~\cite{li2024r}—to assess its ability to tackle realistic and complex tasks. As shown in Table~\ref{tab:benchmark_multi_image_real_world}, even the smallest variant in the InternVL3 family (InternVL3-1B) demonstrates promising performance with a RealWorldQA score of 58.2, an MME-RealWorld score of 46.0, a WildVision win rate of 43.8, and an R-Bench score of 60.4. 
Scaling up the model yields further enhancements across all metrics. Mid-sized variants such as InternVL3-8B and InternVL3-14B continue this positive trend, with InternVL3-8B reporting a RealWorldQA score of 70.8 and an R-Bench score of 74.1. These improvements highlight the effectiveness of scaling, as larger models provide more robust representations and enhanced comprehension capabilities in real-world scenarios.

At the higher end of the scale, the InternVL3-38B and InternVL3-78B models achieve top-tier results among the InternVL3 series. Notably, InternVL3-78B records a RealWorldQA score of 78.0, an MME-RealWorld score of 65.4, a WildVision win rate of 73.6, and an R-Bench score of 77.4. When compared with competitive models, such as GPT-4o~\cite{gpt4v}—which scores 75.4 on RealWorldQA and 80.6 on WildVision—the InternVL3 series exhibits competitive strengths. InternVL3-78B not only surpasses GPT-4o on RealWorldQA and closely matches its R-Bench performance but also considerably outperforms it on MME-RealWorld, indicating an overall robust performance on tasks demanding both perceptual precision and comprehensive understanding.

\begin{table*}[t!]
\centering
\renewcommand{\arraystretch}{0.95}
{\fontsize{7.5}{10}\selectfont 
\setlength\tabcolsep{1.5pt}
\newcommand{\MME}{\makecell{MME\\(sum)}}
\newcommand{\MMB}{\makecell{MMB\\(EN / CN)}}
\newcommand{\MMBV}{\makecell{MMBv1.1\\(EN)}}
\newcommand{\MMVet}{\makecell{MMVet\\(turbo)}}
\newcommand{\MMVetV}{\makecell{MMVetv2\\(0613)}}
\newcommand{\MMStar}{\makecell{MMStar}}
\newcommand{\POPE}{\makecell{POPE\\(avg)}}
\newcommand{\HallB}{\makecell{HallBench\\(avg)}}
\newcommand{\AMBER}{\makecell{AMBER\\(generative / discriminative / overall)}}
\newcommand{\ObjectHalBench}{\makecell{ObjectHal}}
\newcommand{\MMHal}{\makecell{MMHal\\(score)}}
\newcommand{\CRPE}{\makecell{CRPE\\(relation)}}
\begin{tabular}{l|ccccccc|ccccc}
Model Name                                & \MME   & \MMB        & \MMBV & \MMVet & \MMVetV & \MMStar & Overall & \HallB & \MMHal & \CRPE & \POPE & Overall \\
\hline
LLaVA-OneVision-0.5B~\cite{li2024llavaov} & 1438.0 & 61.6 / 55.5 & 59.6  & 32.2   & --      & 37.7    & --      & 27.9   & --     & --    & --    & --      \\
InternVL2-1B~\cite{chen2024far}           & 1794.4 & 65.4 / 60.7 & 61.6  & 32.7   & 36.1    & 45.7    & 51.7    & 34.0   & 2.25   & 57.5  & 87.3  & 45.3    \\
InternVL2.5-1B~\cite{chen2024expanding}   & 1950.5 & 70.7 / 66.3 & 68.4  & 48.8   & 43.2    & 50.1    & 58.9    & 39.0   & 2.49   & 60.9  & 89.9  & 48.1    \\
\rowcolor{gray!15}
InternVL3-1B                              & 1934.4 & 72.6 / 67.9 & 69.9  & 59.5   &  47.5   & 51.5    & 61.9    & 41.4   & 2.59   & 64.0  & 90.7  & 49.7    \\
Qwen2-VL-2B~\cite{wang2024qwen2vl}        & 1872.0 & 74.9 / 73.5 & 72.2  & 49.5   & --      & 48.0    & --      & 41.7   & --     & --    & --    & --      \\
Qwen2.5-VL-3B~\cite{bai2025qwen2}         &   2157 & 79.1 / 78.1 & 77.4  &  61.8  & --      & 55.9    & --      & 46.3   & --     & 73.6  & --    & --      \\
InternVL2-2B~\cite{chen2024far}           & 1876.8 & 73.2 / 70.9 & 70.2  & 39.5   & 39.6    & 50.1    & 58.0    & 37.9   & 2.52   & 66.3  & 88.3  & 48.8    \\
InternVL2.5-2B~\cite{chen2024expanding}   & 2138.2 & 74.7 / 71.9 & 72.2  & 60.8   & 52.3    & 53.7    & 65.3    & 42.6   & 2.94   & 70.2  & 90.6  & 51.6    \\
\rowcolor{gray!15}
InternVL3-2B                              & 2221.2 & 81.1 / 78.4 & 78.6  & 62.2   & 53.9    & 60.7    & 69.8    & 42.5   & 3.26   & 71.5  & 89.6  & 51.7    \\
\hline
Qwen2-VL-7B~\cite{wang2024qwen2vl}        & 2326.8 & 83.0 / 80.5 & 80.7  & 62.0   & --      & 60.7    & --      & 50.6   & 3.40   & 74.4  & 88.1  & 54.1    \\
Qwen2.5-VL-7B~\cite{bai2025qwen2}         & 2347   & 83.5 / 83.4 & 82.6  &  67.1  & --      & 63.9    & --      & 52.9   & --     & 76.4  & --    & --      \\
MiniCPM-V2.6~\cite{yao2024minicpm}        & 2348.4 & 81.5 / 79.3 & 78.0  & 60.0   & --      & 57.5    & --      & 48.1   & 3.60   & 75.2  & 87.3  & 53.6    \\
InternVL2-8B~\cite{chen2024far}           & 2210.3 & 81.7 / 81.2 & 79.5  & 54.2   & 52.3    & 62.0    & 69.2    & 45.2   & 3.33   & 75.8  & 86.9  & 52.8    \\
InternVL2.5-8B~\cite{chen2024expanding}   & 2344.1 & 84.6 / 82.6 & 83.2  & 62.8   & 58.1    & 62.8    & 73.2    & 50.1   & 3.65   & 78.4  & 90.6  & 55.7    \\
\rowcolor{gray!15}
InternVL3-8B                              & 2415.4 & 83.4 / 82.2 & 81.7  & 81.3   & 66.3    & 68.2    & 77.7    & 49.9   & 3.61   & 76.3  & 91.1  & 55.2    \\
\rowcolor{gray!15}
InternVL3-9B                              & 2372.8 & 83.4 / 82.2 & 81.7  & 76.2   & 65.4    & 66.3    & 76.3    & 51.2   & 3.47   & 75.0  & 90.4  & 55.0    \\
\rowcolor{gray!15}
InternVL3-14B                             & 2478.3 & 85.6 / 84.1 & 83.5  & 80.2   & 68.4    & 68.8    & 79.0    & 55.1   & 3.49   & 77.3  & 90.2  & 56.5    \\
\hline
InternVL-Chat-V1.5~\cite{chen2024far}     & 2194.2 & 82.2 / 82.0 & 80.3  & 61.5   & 51.5    & 57.3    & 69.7    & 50.3   & 3.11   & 75.4  & 88.4  & 54.3    \\
InternVL2-26B~\cite{chen2024far}          & 2260.7 & 83.4 / 82.0 & 81.5  & 62.1   & 57.2    & 61.2    & 71.8    & 50.7   & 3.55   & 75.6  & 88.0  & 54.5    \\
InternVL2.5-26B~\cite{chen2024expanding}  & 2373.3 & 85.4 / 85.5 & 84.2  & 65.0   & 60.8    & 66.5    & 75.2    & 55.0   & 3.70   & 79.1  & 90.6  & 57.1    \\
Cambrian-34B~\cite{tong2024cambrian}      & --     & 80.4 / 79.2 & 78.3  & 53.2   & --      & 54.2    & --      & 41.6   & --     & --    & --    & --      \\
InternVL2-40B~\cite{chen2024far}          & 2307.5 & 86.8 / 86.5 & 85.1  & 65.5   & 63.8    & 65.4    & 75.7    & 56.9   & 3.75   & 77.6  & 88.4  & 56.7    \\
InternVL2.5-38B~\cite{chen2024expanding}  & 2455.8 & 86.5 / 86.3 & 85.5  & 68.8   & 62.1    & 67.9    & 77.0    & 56.8   & 3.71   & 78.3  & 90.7  & 57.4    \\
\rowcolor{gray!15}
InternVL3-38B                             & 2523.6 & 87.6 / 86.8 & 86.9 & 83.9   & 69.6    & 71.5    & 81.5    & 57.1   & 3.77   & 77.1  & 90.6  & 57.1    \\
\hline
GPT-4V~\cite{gpt4v}                       & 1926.6 & 81.0 / 80.2 & 80.0  & 67.5   & 66.3    & 56.0    & 70.7    & 46.5   & --     & --    & --    & --      \\
GPT-4o-20240513~\cite{gpt4v}              & --     & 83.4 / 82.1 & 83.1  & 69.1   & 71.0    & 64.7    & --      & 55.0   & 4.00   & 76.6  & 86.9  & 55.6    \\
Claude-3-Opus~\cite{claude3series2024}    & 1586.8 & 63.3 / 59.2 & 60.1  & 51.7   & 55.8    & 45.7    & 55.5    & 37.8   & --     & --    & --    & --      \\
Claude-3.5-Sonnet~\cite{claude3series2024}& --     & 82.6 / 83.5 & 80.9  & 70.1   & 71.8    & 65.1    & --      & 55.5   & --     & --    & --    & --      \\
Gemini-1.5-Pro~\cite{reid2024gemini1_5}   & --     & 73.9 / 73.8 & 74.6  & 64.0   & 66.9    & 59.1    & --      & 45.6   & --     & --    & --    & --      \\
LLaVA-OneVision-72B~\cite{li2024llavaov}  & 2261.0 & 85.8 / 85.3 & 85.0  & 60.6   & --      & 65.8    & --      & 49.0   & --     & --    & --    & --      \\
Qwen2-VL-72B~\cite{wang2024qwen2vl}       & 2482.7 & 86.5 / 86.6 & 85.9  & 74.0   & 66.9    & 68.3    & 78.7    & 58.1   & --     & --    & --    & --      \\
Qwen2.5-VL-72B~\cite{bai2025qwen2}        & 2448.0 & 88.6 / 87.9 & 88.4  & 76.2   & --      & 70.8    & --      & 55.2   & --     & 79.2  & --    & --      \\
InternVL2-Llama3-76B~\cite{chen2024far}   & 2414.7 & 86.5 / 86.3 & 85.5  & 65.7   & 68.4    & 67.4    & 77.2    & 55.2   & 3.83   & 77.6  & 89.0  & 56.4    \\
InternVL2.5-78B~\cite{chen2024expanding}  & 2494.5 & 88.3 / 88.5 & 87.4  & 72.3   & 65.5    & 69.5    & 79.2    & 57.4   & 3.89   & 78.8  & 90.8  & 57.7    \\
\rowcolor{gray!15}
InternVL3-78B                             & 2549.8 & 89.0 / 88.7 & 87.7  & 81.3   & 70.0    & 72.5    & 82.0    & 59.1   & 3.85   & 79.2  & 90.3  & 58.1    \\
\end{tabular}
}

\caption{\textbf{Comparison of comprehensive multimodal understanding and hallucination performance.}
Comprehensive multimodal benchmarks include MME~\cite{fu2023mme}, MMBench series~\cite{liu2023mmbench}, MMVet series~\cite{yu2023mmvet, yu2024mmvetv2}, and MMStar~\cite{chen2024mmstar}.
Hallucination benchmarks encompass HallusionBench~\cite{guan2023hallusionbench}, MMHal~\cite{sun2023aligning}, CRPE~\cite{wang2024allseeingv2}, and POPE~\cite{li2023pope}.
Part of the results are sourced from the benchmark papers and the OpenCompass leaderboard~\cite{opencompass2023}.
}
\label{tab:benchmark_multimodal_hallucination}
\end{table*}

\subsection{Comprehensive Multimodal Evaluation}

The comprehensive multimodal evaluation is based on established benchmarks including MME~\cite{fu2023mme}, MMBench (evaluating both English and Chinese tasks)~\cite{liu2023mmbench}, MMBench v1.1 (English)~\cite{liu2023mmbench}, MMVet~\cite{yu2023mmvet}, MMVet v2~\cite{yu2024mmvetv2}, and MMStar~\cite{chen2024mmstar}, as summarized in Table~\ref{tab:benchmark_multimodal_hallucination}.
Specifically, InternVL3-1B achieves an MMBench score of 72.6/67.9 (English/Chinese) and improves the MMBench v1.1 score to 69.9, compared to the InternVL2.5-1B baseline (70.7/66.3 and 68.4, respectively).
The improvements become more pronounced at the 2B scale, where InternVL3-2B records an MME of 2221.2 and reaches an MMBench performance of 81.1/78.4, along with an MMBench v1.1 score of 78.6. 

At larger scales, InternVL3 models consistently demonstrate superior performance. For example, the InternVL3-8B model achieves an MME of 2415.4, while the InternVL3-38B and InternVL3-78B models record MME scores of 2523.6 and 2549.8, respectively. The corresponding MMBench and MMBench v1.1 scores also show steady improvements, with InternVL3-78B attaining 89.0/88.7 for English/Chinese and 87.7 for English-only tasks. When compared with other competitive models, such as Qwen2-VL-72B and Qwen2.5-VL-72B, the InternVL3 series—especially the 78B variant—offers a consistent performance advantage on the multimodal understanding benchmarks.

\subsection{Multimodal Hallucination Evaluation}

\label{sec:hall_datasets}

We evaluate InternVL’s propensity for hallucinations on four established benchmarks—HallusionBench~\cite{guan2023hallusionbench}, MMHal-Bench~\cite{sun2023aligning}, CRPE~\cite{wang2024allseeingv2}, and POPE~\cite{li2023pope}—as detailed in Table~\ref{tab:benchmark_multimodal_hallucination}. In comparison with previous InternVL series, the new InternVL3 models demonstrate overall competitive performance across varying scales, while providing consistent improvements in handling multimodal hallucination challenges.
In the small-parameter regime, InternVL3-1B attains a HallusionBench score of 41.4, representing an appreciable gain over the InternVL2.5-1B baseline, which scored 39.0.  Similarly, the 2B variant of InternVL3 shows a comparable HallusionBench performance (42.5) to its InternVL2.5 counterpart (42.6), while registering a modest improvement in CRPE performance (71.5 \vs~70.2). 

In the large-scale setting, InternVL3-38B and InternVL3-78B are particularly noteworthy. InternVL3-38B obtains a HallusionBench score of 57.1, while InternVL3-78B reaches 59.1, accompanied by a CRPE improvement to 79.2. These figures position the InternVL3 series as competitive with leading closed- and open-source models such as GPT-4o and the Qwen2.5-VL series. Despite these advancements, minor declines on certain benchmarks, such as MMHal, indicate that although the InternVL3 series has made overall progress, optimizing data and training strategies to achieve more consistent improvements remains an important direction for future work.

\subsection{Visual Grounding}

\begin{table*}[t] 
\centering 
\renewcommand{\arraystretch}{0.95}
\setlength\tabcolsep{8pt}
{\fontsize{8}{10}\selectfont 
\begin{tabular}{l|ccc|ccc|cc|c}
\multirow{2}{*}{{Model Name}}           & \multicolumn{3}{c|}{{RefCOCO}} & \multicolumn{3}{c|}{{RefCOCO+}} & \multicolumn{2}{c|}{{RefCOCOg}} & \multirow{2}{*}{{Overall}} \\
                                        & val    & test-A    & test-B    & val    & test-A    & test-B     & val     & test                 &         \\
\hline
Grounding-DINO-L~\citep{grounding_dino} & 90.6   & 93.2      & 88.2      & 82.8   & 89.0      & 75.9       & 86.1    & 87.0                 & 86.6    \\
UNINEXT-H~\citep{uninext}               & 92.6   & 94.3      & 91.5      & 85.2   & 89.6      & 79.8       & 88.7    & 89.4                 & 88.9    \\
ONE-PEACE~\citep{one-peace}             & 92.6   & 94.2      & 89.3      & 88.8   & 92.2      & 83.2       & 89.2    & 89.3                 & 89.8    \\
\hline
Qwen2.5-VL-3B~\cite{bai2025qwen2}       & 89.1   & 91.7      & 84.0      & 82.4   & 88.0      & 74.1       & 85.2    & 85.7                 & 85.0    \\
\rowcolor{gray!15}
InternVL3-1B                             & 85.8   & 90.1      & 81.7      & 76.6   &	84.1	  & 69.2	   & 82.8    & 82.6                 & 81.6    \\
\rowcolor{gray!15}
InternVL3-2B                             & 89.8   & 92.6	     & 86.4      & 84.0	  & 89.2      & 76.5       & 87.6    & 87.2                 & 86.7    \\
Shikra-7B~\citep{chen2023shikra}        & 87.0   & 90.6      & 80.2      & 81.6   & 87.4      & 72.1       & 82.3    & 82.2                 & 82.9    \\
Ferret-v2-13B~\citep{ferretv2}          & 92.6   & 95.0      & 88.9      & 87.4   & 92.1      & 81.4       & 89.4    & 90.0                 & 89.6    \\
CogVLM-Grounding~\citep{wang2023cogvlm} 
                                        & 92.8   & 94.8      & 89.0      & 88.7   & 92.9      & 83.4       & 89.8    & 90.8                 & 90.3    \\
MM1.5~\cite{zhang2024mm1_5}             & --     & 92.5      & 86.7      & --     & 88.7      & 77.8       & --      & 87.1                 & --      \\
Qwen2-VL-7B~\cite{wang2024qwen2vl}      & 91.7   & 93.6      & 87.3      & 85.8   & 90.5      & 79.5       & 87.3    & 87.8                 & 87.9    \\
Qwen2.5-VL-7B~\cite{bai2025qwen2}       & 90.0   & 92.5      & 85.4      & 84.2   & 89.1      & 76.9       & 87.2    & 87.2                 & 86.6    \\
TextHawk2~\citep{yu2024texthawk2}       & 91.9   & 93.0      & 87.6      & 86.2   & 90.0      & 80.4       & 88.2    & 88.1                 & 88.2    \\
InternVL2-8B~\cite{chen2024far}         & 87.1   & 91.1      & 80.7      & 79.8   & 87.9      & 71.4       & 82.7    & 82.7                 & 82.9    \\

InternVL2.5-8B~\cite{chen2024expanding} & 90.3   & 94.5      & 85.9      & 85.2   & 91.5      & 78.8       & 86.7    & 87.6                 & 87.6    \\
\rowcolor{gray!15}
InternVL3-8B                            & 92.5   & 94.6      & 88.0      & 88.2   & 92.5      & 81.8       & 89.6    & 90.0                 & 89.6    \\
\rowcolor{gray!15}
InternVL3-9B                            & 91.8  & 93.2     & 86.6     & 86.4  & 91.0     & 79.9      & 88.0   & 88.5                & 88.2    \\  
\rowcolor{gray!15}
InternVL3-14B                           & 92.0  & 94.4     & 87.8     & 87.4  & 92.1     & 81.5      & 88.6   & 89.3                 & 89.1    \\  
\hline

Qwen2-VL-72B~\cite{wang2024qwen2vl}     & 93.2   & 95.3      & 90.7      & 90.1   & 93.8      & 85.6       & 89.9    & 90.4                 & 91.1    \\
Qwen2.5-VL-72B~\cite{bai2025qwen2}      & 92.7   & 94.6      & 89.7      & 88.9   & 92.2      & 83.7       & 89.9    & 90.3                 & 90.3    \\
InternVL2-Llama3-76B~\cite{chen2024far} & 92.2   & 94.8      & 88.4      & 88.8   & 93.1      & 82.8       & 89.5    & 90.3                 & 90.0    \\
InternVL2.5-78B~\cite{chen2024expanding}& 93.7   & 95.6      & 92.5      & 90.4   & 94.7      & 86.9       & 92.7    & 92.2                 & 92.3    \\
\rowcolor{gray!15}
InternVL3-38B                           & 93.2   & 95.1      & 90.2      & 89.8   & 93.2      & 85.2       & 91.4    & 91.5                 & 91.2    \\
\rowcolor{gray!15}  
InternVL3-78B                           & 93.4   & 95.4      & 90.3      & 90.1   & 93.8      & 85.3       & 91.5    & 91.5                 & 91.4    \\
\end{tabular}
}
\caption{\textbf{Comparison of visual grounding performance.}
We evaluate InternVL's visual grounding capability on RefCOCO, RefCOCO+, and RefCOCOg datasets~\cite{kazemzadeh2014referitgame, mao2016generation}. Parts of the results are collected from \cite{wang2024qwen2vl}.
} 
\label{tab:benchmark-grounding}
\end{table*}

We evaluate InternVL’s visual grounding capability on the RefCOCO~\cite{kazemzadeh2014referitgame}, RefCOCO+\cite{kazemzadeh2014referitgame}, and RefCOCOg\cite{mao2016generation} datasets, where the model is tasked with accurately localizing target objects in images from given textual descriptions. Table~\ref{tab:benchmark-grounding} shows a comprehensive comparison across various models, including several specialized grounding models as well as multiple MLLLMs.

Among the smaller-scale models, we observe that while Qwen2.5-VL-3B achieves an average score of 85.0, the InternVL3-1B and InternVL3-2B models yield average scores of 81.6 and 86.7, respectively.
Notably, when scaling up, the InternVL3 series exhibits promising improvements. InternVL3-8B, InternVL3-9B, and InternVL3-14B yield average scores around 88.2–89.6, reflecting a consistent trend of performance gains as the model size increases. However, when reaching larger scales, the performance gains appear to plateau. For instance, InternVL2.5-78B reaches an average score of 92.3, and InternVL3-78B only shows a score of 91.4.
We speculate that this is because InternVL3’s training data expansion does not include additional grounding-specific data and the relative reduction in grounding-targeted data could have restricted the localization capabilities.

\subsection{Multimodal Multilingual Understanding}

\begin{table*}[t!]
\scriptsize
\centering
\renewcommand{\arraystretch}{0.95}
{\fontsize{8}{10}\selectfont 
\setlength\tabcolsep{4.2pt}
\newcommand{\MMMB}{\makecell{MMMB}}
\newcommand{\MultiMMB}{\makecell{Multilingual MMBench}}
\begin{tabular}{l|cccccc|cccccc|c|c}
\multirow{2}{*}{Model Name}               & \multicolumn{6}{c|}{\MMMB}              & \multicolumn{6}{c|}{\MultiMMB}          & MTVQA   & \multirow{2}{*}{Overall} \\
                                          & en   & zh   & pt   & ar   & tr   & ru   & en   & zh   & pt   & ar   & tr   & ru   & (avg)   &                          \\
\hline
InternVL2-1B~\cite{chen2024far}           & 73.2 & 67.4 & 55.5 & 53.5 & 43.8 & 55.2 & 67.9 & 61.2 & 50.8 & 43.3 & 31.8 & 52.7 & 12.6    & 40.7                     \\
InternVL2.5-1B~\cite{chen2024expanding}   & 78.8 & 70.2 & 61.5 & 55.0 & 45.3 & 61.1 & 72.5 & 64.7 & 57.0 & 43.0 & 37.8 & 53.2 & 21.4    & 46.0                     \\
\rowcolor{gray!15}
InternVL3-1B                              & 79.4 & 70.1 & 62.3 & 58.0 &	47.6 & 61.9 & 72.6 & 66.2 & 62.3 & 48.0 & 39.5 & 60.3 & 22.2    & 47.9                     \\
Qwen2-VL-2B~\cite{wang2024qwen2vl}        & 78.3 & 74.2 & 72.6 & 68.3 & 61.8 & 72.8 & 72.1 & 71.1 & 69.9 & 61.1 & 54.4 & 69.3 & 20.0    & 52.6                     \\
Qwen2.5-VL-3B~\cite{bai2025qwen2}         & --   & --   & --   & --   & --   & --   & --   & --   & --   & --   & --   & --   & 24.8    & --                       \\
InternVL2-2B~\cite{chen2024far}           & 79.4 & 71.6 & 54.0 & 43.5 & 46.4 & 48.1 & 73.8 & 69.6 & 51.4 & 29.8 & 31.3 & 42.3 & 10.9    & 39.3                     \\
InternVL2.5-2B~\cite{chen2024expanding}   & 81.4 & 74.4 & 58.2 & 48.3 & 46.4 & 53.2 & 76.5 & 71.6 & 55.9 & 37.3 & 33.9 & 44.8 & 21.8    & 45.2                     \\
\rowcolor{gray!15}
InternVL3-2B                              & 81.9 & 78.3 & 75.4 & 68.6 & 62.9 & 74.6 & 81.3 & 77.8 & 75.9 & 66.4 & 59.5 & 70.7 & 26.7    & 57.4                     \\
\hline
mPLUG-Owl2~\cite{ye2023mplug2}            & 67.3 & 61.0 & 59.7 & 45.8 & 45.4 & 62.6 & 66.2 & 59.4 & 58.2 & 37.9 & 47.7 & 60.4 & --      & --                       \\
Qwen2-VL-7B~\cite{wang2024qwen2vl}        & 83.9 & 82.4 & 81.2 & 79.0 & 74.7 & 82.4 & 81.8 & 81.6 & 79.1 & 75.6 & 74.5 & 79.3 & 25.6    & 61.6                     \\
Qwen2.5-VL-7B~\cite{bai2025qwen2}         & --   & --   & --   & --   & --   & --   & --   & --   & --   & --   & --   & --   & 29.2    & --                       \\
InternVL2-8B~\cite{chen2024far}           & 83.4 & 81.5 & 76.1 & 66.3 & 69.2 & 75.7 & 82.9 & 81.8 & 76.0 & 60.5 & 66.0 & 74.4 & 20.9    & 56.6                     \\ 
InternVL2.5-8B~\cite{chen2024expanding}   & 84.3 & 83.1 & 78.6 & 69.3 & 71.5 & 79.5 & 83.8 & 83.2 & 79.4 & 64.3 & 67.8 & 77.3 & 27.6    & 60.4                     \\
\rowcolor{gray!15}
InternVL3-8B                              & 85.1 & 83.1 & 82.5 & 81.6 & 76.2 & 83.4 & 85.5 & 85.6 &	83.2 & 79.2 & 75.9 & 82.6 & 30.2    & 64.7                     \\
\rowcolor{gray!15}
InternVL3-9B                              & 84.8 & 83.7 & 80.6 & 69.9 & 68.5 & 80.8 & 86.5 & 85.2 & 79.1 & 64.3 & 68.3 & 79.1 & 27.1    & 60.7                     \\
\rowcolor{gray!15}
InternVL3-14B                             & 85.7 & 84.7 & 83.1 & 83.7 &	79.3 & 83.6 & 86.7 & 85.8 & 83.2 & 81.1 & 80.7 & 83.8 & 31.6    & 66.2                     \\
\hline
InternVL-Chat-V1.5~\cite{chen2024far}     & 82.6 & 80.8 & 76.3 & 65.2 & 68.6 & 74.0 & 81.1 & 80.2 & 76.9 & 56.2 & 66.7 & 71.0 & 20.5    & 55.7                     \\
InternVL2-26B~\cite{chen2024far}          & 83.8 & 81.7 & 78.0 & 68.8 & 69.3 & 76.3 & 82.7 & 81.8 & 77.8 & 61.9 & 69.6 & 74.4 & 17.7    & 56.2                     \\
InternVL2.5-26B~\cite{chen2024expanding}  & 86.2 & 83.8 & 81.6 & 73.3 & 73.7 & 82.8 & 86.1 & 85.5 & 80.7 & 67.5 & 75.0 & 79.6 & 28.5    & 62.6                     \\
InternVL2-40B~\cite{chen2024far}          & 85.3 & 84.1 & 81.1 & 70.3 & 74.2 & 81.4 & 86.2 & 85.8 & 82.8 & 64.0 & 74.2 & 81.8 & 20.6    & 59.7                     \\
InternVL2.5-38B~\cite{chen2024expanding}  & 86.4 & 85.1 & 84.1 & 84.3 & 82.8 & 84.9 & 87.5 & 88.6 & 85.3 & 84.5 & 84.0 & 85.9 & 31.7    & 67.4                     \\
\rowcolor{gray!15}
InternVL3-38B                             & 86.7 & 85.6 & 84.5 & 84.8 &	82.6 & 85.1 & 89.0 & 89.3 & 87.1 & 84.6 & 84.3 & 87.4 & 32.4    & 68.1                     \\
\hline
GPT-4V~\cite{gpt4v}                       & 75.0 & 74.2 & 71.5 & 73.5 & 69.0 & 73.1 & 77.6 & 74.4 & 72.5 & 72.3 & 70.5 & 74.8 & 22.0    & 56.1                     \\
GPT-4o~\cite{gpt4v}                       & --   & --   & --   & --   & --   & --   & --   & --   & --   & --   & --   & --   & 27.8    & --                       \\
Gemini-1.0-Pro~\cite{team2023gemini}      & 75.0 & 71.9 & 70.6 & 69.9 & 69.6 & 72.7 & 73.6 & 72.1 & 70.3 & 61.1 & 69.8 & 70.5 & --      & --                       \\
Qwen2-VL-72B~\cite{wang2024qwen2vl}       & 86.8 & 85.3 & 85.2 & 84.8 & 84.2 & 85.3 & 86.9 & 87.2 & 85.8 & 83.5 & 84.4 & 85.3 & 30.9    & 67.2                     \\
Qwen2.5-VL-72B~\cite{bai2025qwen2}        & --   & --   & --   & --   & --   & --   & --   & --   & --   & --   & --   & --   & 31.7    & --                       \\
InternVL2-Llama3-76B~\cite{chen2024far}   & 85.3 & 85.1 & 82.8 & 82.8 & 83.0 & 83.7 & 87.8 & 87.3 &	85.9 & 83.1 & 85.0 & 85.7 & 22.0    & 63.9                     \\
InternVL2.5-78B~\cite{chen2024expanding}  & 86.3 & 85.6 & 85.1 & 84.8 & 83.1 & 85.4 & 90.0 & 89.7 & 87.4 & 83.3 & 84.9 & 86.3 & 31.9    & 68.0                     \\
\rowcolor{gray!15}
InternVL3-78B                             & 87.2 & 86.6 & 85.5 & 86.5 & 84.6 & 86.1 & 89.4 & 90.3 &	88.7 & 86.1 & 86.6 & 88.1 & 32.5    & 68.9                     \\
\end{tabular}
}
\caption{\textbf{Comparison of multimodal multilingual performance. }
We evaluate multilingual capabilities across 3 benchmarks, including MMMB~\cite{sun2024parrot}, Multilingual MMBench~\cite{sun2024parrot} and MTVQA~\cite{tang2024mtvqa}.
The languages evaluated are English (en), Chinese (zh), Portuguese (pt), Arabic (ar), Turkish (tr), and Russian (ru).
}
\label{tab:benchmark_multilingual}
\end{table*}

\label{sec:multilingual_datasets}

We assess InternVL's multimodal multilingual understanding capabilities using benchmarks—MMMB, Multilingual MMBench~\cite{sun2024parrot}, and MTVQA~\cite{tang2024mtvqa}—as shown in Table~\ref{tab:benchmark_multilingual}. The InternVL3 series demonstrates consistent improvements in multilingual performance compared to previous predecessors. For example, the lightweight InternVL3-1B already shows a modest improvement over InternVL2.5-1B, while the larger-scale variants, such as InternVL3-38B and InternVL3-78B, achieve significantly higher average scores across all three benchmarks.

Comparisons with other leading models further highlight the effectiveness of the InternVL3 series. Notably, the InternVL3 variants achieve performance that is competitive with or superior to models such as Qwen2-VL-72B~\cite{wang2024qwen2vl} and Qwen2.5-VL-72B~\cite{bai2025qwen2}. Overall, the enhanced performance of the InternVL3 series across MMMB, Multilingual MMBench, and MTVQA underscores the promise of our approach in advancing global multimodal applications.

\subsection{Video Understanding}

\begin{table*}[t!]
\centering
\renewcommand{\arraystretch}{0.95}
\setlength\tabcolsep{3pt}
\newcommand{\VMME}{\makecell{Video-MME\\(wo / w sub)}}
\newcommand{\MVB}{\makecell{MVBench}}
\newcommand{\MMBV}{\makecell{MMBench-Video\\(val)}}
\newcommand{\MLVU}{\makecell{MLVU\\(M-Avg)}}
\newcommand{\LVB}{\makecell{LongVideoBench\\(val total)}}
\newcommand{\CG}{\makecell{CG-Bench\\(long / clue acc.)}}
\newcommand{\lsp}{--\ \ \ \ \ \ }
\newcommand{\rsp}{\ \ \ \ \ --~}
{\fontsize{8}{10}\selectfont 
\begin{tabular}{l|cccccc|c}
Model Name                                 &    \VMME    & \MVB & \MMBV & \MLVU & \LVB  & \CG         & Overall \\
\hline
InternVL2-1B~\cite{chen2024far}            & 42.9 / 45.4 & 57.5 & 1.14  & 51.6  & 43.3  & --          & --      \\
InternVL2.5-1B~\cite{chen2024expanding}    & 50.3 / 52.3 & 64.3 & 1.36  & 57.3  & 47.9  & --          & --      \\
\rowcolor{gray!15}
InternVL3-1B                               & 51.0 / 53.0 & 63.1 &  1.3  & 53.0  & 48.1  & 24.8 / 39.1 & 46.9    \\
Qwen2-VL-2B~\cite{wang2024qwen2vl}         & 55.6 / 60.4 & 63.2 & --    & --    & --    & --          & --      \\
Qwen2.5-VL-3B~\cite{bai2025qwen2_5}        & 61.5 / 67.6 & 67.0 & 1.63  & 68.2  & 43.3  & --          & --      \\
InternVL2-2B~\cite{chen2024far}            & 46.2 / 49.1 & 60.2 & 1.30  & 54.3  & 46.0  & --          & --      \\
InternVL2.5-2B~\cite{chen2024expanding}    & 51.9 / 54.1 & 68.8 & 1.44  & 61.4  & 52.0  & --          & --      \\
\rowcolor{gray!15}
InternVL3-2B                               & 58.9 / 61.4 & 70.4 & 1.42  & 64.2  & 55.4  & 30.8 / 50.7 & 54.9    \\
\hline
VideoChat2-HD~\cite{li2023videochat}       & 45.3 / 55.7 & 62.3 & 1.22  & 47.9  & --    & --          & --      \\
MiniCPM-V-2.6~\cite{yao2024minicpm}        & 60.9 / 63.6 & --   & 1.70  & --    & 54.9  & --          & --      \\
LLaVA-OneVision-7B~\cite{li2024llavaov}    & 58.2 / \lsp & 56.7 & --    & --    & --    & --          & --      \\
Qwen2-VL-7B~\cite{wang2024qwen2vl}         & 63.3 / 69.0 & 67.0 & 1.44  & --    & 55.6  & --          & --      \\
Qwen2.5-VL-7B~\cite{bai2025qwen2_5}        & 65.1 / 71.6 & 69.6 & 1.79  & 70.2  & 45.3  & --          & --      \\
InternVL2-8B~\cite{chen2024far}            & 56.3 / 59.3 & 65.8 & 1.57  & 64.0  & 54.6  & --          & --      \\
InternVL2.5-8B~\cite{chen2024expanding}    & 64.2 / 66.9 & 72.0 & 1.68  & 68.9  & 60.0  & --          & --      \\
\rowcolor{gray!15}
InternVL3-8B                               & 66.3 / 68.9 & 75.4 & 1.69  & 71.4  & 58.8  & 38.6 / 55.2 & 61.4    \\
\rowcolor{gray!15}
InternVL3-9B                               & 66.7 / 68.9 & 74.3 & 1.69  & 70.8  & 62.5  & 41.1 / 58.0 & 62.3    \\
\rowcolor{gray!15}
InternVL3-14B                              & 70.4 / 73.0 & 76.6 & 1.73  & 73.3  & 63.9  & 44.1 / 60.6 & 64.9    \\
\hline
InternVL2-26B~\cite{chen2024far}           & 57.0 / 60.2 & 67.5 & 1.67  & 64.2  & 56.1  & --          & --      \\
InternVL2.5-26B                            & 66.9 / 69.2 & 75.2 & 1.86  & 72.3  & 59.9  & --          & --      \\
Oryx-1.5-32B~\cite{liu2024oryx}            & 67.3 / 74.9 & 70.1 & 1.52  & 72.3  & --    & --          & --      \\  
Qwen2.5-VL-32B~\cite{bai2025qwen2_5}       & 70.5 / 77.9 & --   & 1.93  & --    & --    & --          & --      \\
VILA-1.5-40B~\cite{lin2024vila}            & 60.1 / 61.1 & --   & 1.61  & 56.7  & --    & --          & --      \\
InternVL2-40B~\cite{chen2024far}           & 66.1 / 68.6 & 72.0 & 1.78  & 71.0  & 60.6  & --          & --      \\
InternVL2.5-38B~\cite{chen2024expanding}   & 70.7 / 73.1 & 74.4 & 1.82  & 75.3  & 63.3  & --          & --      \\
\rowcolor{gray!15}
InternVL3-38B                              & 72.7 / 75.0 & 76.9 & 1.81  & 77.8  & 67.3  & 46.9 / 62.8 & 67.5    \\
\hline
GPT-4V/4T~\cite{openai2023gpt4}            & 59.9 / 63.3 & 43.7 & 1.53  & 49.2  & 59.1  & --          & --      \\
GPT-4o-20240513~\cite{gpt4v}               & 71.9 / 77.2 & --   & 1.63  & 64.6  & 66.7  & --          & --      \\
GPT-4o-20240806~\cite{gpt4v}               & --          & --   & 1.87  & --    & --    & 41.8 / 58.3 & --      \\
Gemini-1.5-Pro~\cite{reid2024gemini1_5}    & 75.0 / 81.3 & --   & 1.30  & --    & 64.0  & 40.1 / 56.4 & --      \\
VideoLLaMA2-72B~\cite{cheng2024videollama2}& 61.4 / 63.1 & 62.0 & --    & --    & --    & --          & --      \\
LLaVA-OneVision-72B~\cite{li2024llavaov}   & 66.2 / 69.5 & 59.4 & --    & 66.4  & 61.3  & --          & --      \\
Qwen2-VL-72B~\cite{wang2024qwen2vl}        & 71.2 / 77.8 & 73.6 & 1.70  & --    & --    & 41.3 / 56.2 & --      \\
Qwen2.5-VL-72B~\cite{bai2025qwen2_5}       & 73.3 / 79.1 & 70.4 & 2.02  & 74.6  & 60.7  & --          & --      \\
InternVL2-Llama3-76B~\cite{chen2024far}    & 64.7 / 67.8 & 69.6 & 1.71  & 69.9  & 61.1  & --          & --      \\
InternVL2.5-78B~\cite{chen2024expanding}   & 72.1 / 74.0 & 76.4 & 1.97  & 75.7  & 63.6  & 42.2 / 58.5 & 66.0    \\ 
\rowcolor{gray!15}
InternVL3-78B                              & 72.7 / 75.7 & 78.7 & 1.81  & 79.5  & 65.7  & 48.4 / 65.3 & 68.3    \\ 
\end{tabular}
}
\caption{\textbf{Comparison of video understanding performance.}
We evaluate InternVL's video understanding capabilities across 6 benchmarks.
For Video-MME~\cite{fu2024video}, MMBench-Video~\cite{fang2024mmbench}, MLVU~\cite{MLVU}, and LongVideoBench~\cite{wu2024longvideobench}, we test with four different settings: 16, 32, 48, and 64 frames, and report the maximum results. For MVBench~\cite{li2024mvbench}, we conduct testing using 16 frames. For CG-Bench~\cite{anonymous2024cgbench}, we use 32 frames.
}
\label{tab:benchmark_video}
\end{table*}

Video understanding is essential for evaluating how well MLLMs capture temporal and multimodal cues in complex video content. In this work, we assess the InternVL3 series on six established benchmarks—Video-MME~\cite{fu2024video}, MVBench~\cite{li2024mvbench}, MMBench-Video~\cite{fang2024mmbench}, MLVU~\cite{MLVU}, LongVideoBench~\cite{wu2024longvideobench}, and CG-Bench~\cite{anonymous2024cgbench}, as detailed in Table~\ref{tab:benchmark_video}.

Overall, the InternVL3 models demonstrate clear performance improvements and a strong scalability trend over their predecessors. As the model capacity increases, the performance gains become more pronounced. For instance, InternVL3-2B records higher Video-MME scores (58.9/61.4) and improved MVBench and MLVU performance compared to the earlier 2B variants. 

The scaling behavior of the InternVL3 series is further evident in the larger models. InternVL3-14B attains a Video-MME score of 70.4/73.0, while InternVL3-38B and InternVL3-78B push these metrics even higher, reaching scores of 72.7/75.0 and 72.7/75.7, respectively. Additionally, the inclusion of CG-Bench evaluations for the InternVL3 series provides further insight into long-range video reasoning, with performance steadily improving as model size increases—for example, InternVL3-78B attains 48.4/65.3 on CG-Bench.

When compared with other open-source models, the InternVL3 series demonstrates competitive advantages. For instance, while Qwen2.5-VL models achieve impressive results (with Qwen2.5-VL-72B scoring 73.3/79.1 on Video-MME), the InternVL3 series tends to outperform them in other metrics, such as MVBench and MLVU. Similarly, while closed-source systems like Gemini-1.5-Pro sometimes yield superior results on select benchmarks (\eg, Video-MME), the overall performance of InternVL3, especially at larger scales, is highly competitive.

\subsection{GUI Grounding}
\label{sec:gui_grounding}
\begin{table*}[h]
\centering
\small
\footnotesize
{\fontsize{8}{10}\selectfont 
\setlength\tabcolsep{4pt}
\begin{tabular}{l|ccc|ccc|ccc}
\toprule
\textbf{Method} & GPT-4o & Gemini 2.0 & Claude & Aguvis-72B & Qwen2.5-VL-72B & UI-TARS-72B & InternVL3-8B & -38B & -72B\\ 
\midrule
ScreenSpot & 18.1 & 84.0 & 83.0 & \textbf{89.2} & 87.1 & 88.4 & 79.5 & 85.6 & 88.7 \\ 
ScreenSpot-V2 & $-$ & $-$ & $-$ & $-$ & $-$ & 90.3 & 81.4 & 88.3 & \textbf{90.9} \\  
\bottomrule
\end{tabular}
}
\caption{\textbf{Performance of InternVL3 and other models on GUI grounding benchmarks.}}
\label{tab:gui_grounding}
\end{table*}

GUI grounding requires precise localization and understanding of interface elements, which is critical for applications like automated UI testing and assistive technologies.
In Table~\ref{tab:gui_grounding}, we report the performance on GUI grounding benchmarks, comparing InternVL3 with state-of-the-art multimodal and GUI-specific models. The results demonstrate that InternVL3 achieves competitive performance across different scales. On ScreenSpot~\cite{cheng2024seeclick}, InternVL3-72B achieves 88.7\% accuracy, slightly outperforming UI-TARS-72B~\cite{qin2025ui} (88.4\%) and Qwen2.5-VL-72B (87.1\%), while Aguvis-72B~\cite{xu2024aguvis} leads with 89.2\%. Notably, InternVL3-38B (85.6\%) surpasses GPT-4o (18.1\%) and Gemini 2.0 (84.0\%) by a significant margin.

For the more challenging ScreenSpot-V2~\cite{wu2024atlas} benchmark, InternVL3 exhibits strong scaling behavior: InternVL3-72B achieves 90.9\%, outperforming UI-TARS-72B (90.3\%). The 8B variant (81.4\%) already surpasses UI-TARS-72B, while the 38B model (88.3\%) further closes the gap to the 72B version. These results highlight InternVL3's robustness in GUI understanding tasks, particularly in handling complex screen layouts and dynamic interfaces. The performance improvements with model scale suggest that larger architectures better capture the fine-grained visual-textual alignments required for precise GUI grounding.
 The superior performance of the InternVL3 models highlights their robustness in interpreting complex visual layouts. Future work will explore extending these capabilities to more dynamic and interactive GUI environments.

\subsection{Spatial Reasoning}
Spatial reasoning involves constructing a mental representation of a three-dimensional environment from visual inputs—a capability that is vital for applications such as autonomous driving. Table~\ref{tab:vis_benchmark} reports the performance results on the Visual-Spatial Intelligence Benchmark (VSI-Bench)~\cite{yang2024think}, where InternVL3 is compared against other state-of-the-art MLLMs. The results clearly indicate that InternVL3 outperforms its competitors in spatial reasoning tasks.
In particular, the InternVL3-8B variant achieves a score of 42.1, leading all open-source MLLMs in the benchmark. Moreover, the InternVL3-38B and InternVL3-78B variants score 48.9 and 48.4, respectively—both superior to proprietary models such as GPT-4o, Gemini-1.5 Flash, and Gemini-1.5 Pro.

Furthermore, InternVL3 exhibits exceptional performance in several sub-category tasks within the benchmark. It attains a score of 71.2 in object counting, 53.7 in absolute distance estimation, 55.9 in relative distance estimation, and 54.5 in appearance order prediction, demonstrating its robust spatial reasoning capabilities. These promising results underscore the potential of InternVL3 for advancing 3D scene understanding, and future work will explore its integration into various downstream applications.
\begin{table*}[t!]
\centering
\renewcommand{\arraystretch}{0.95}
\setlength\tabcolsep{3pt}
{\fontsize{8}{10}\selectfont 

\begin{tabular}{l|cccccccc|c}

Model Name                                        & Obj.count & Abs.Dist.& Obj.size & Room Size & Rel.Dist. & Rel.Dir.& Route Plan & Appr.Order &   Overall\\
\hline
GPT-4o~\cite{gpt4v}   & 46.2  & 5.3   & 43.8      & 38.2    & 37.0   & 41.3     & 31.5    & 28.5 &  34.0   \\
Gemini-1.5 Pro~\cite{reid2024gemini1_5}  & 56.2  & 30.9   & 64.1      & 43.6    & 51.3   & 46.3     & 36.0    & 34.6  &  45.4  \\
\hline
VILA-1.5-8B~\cite{lin2024vila}  & 17.4  & 21.8   & 50.3      & 18.8    & 32.1   & 34.8     & 31.0    & 24.8  &  28.9  \\
LongVA-7B~\cite{longva}  & 38.0  & 16.6   & 38.9      & 22.2   & 33.1   & 43.3     & 25.4    & 15.7  &  29.2  \\
LLaVA-NeXT-Video-7B~\cite{zhang2024llavanext}  & 48.5  & 14.0   & 47.8      & 24.2    & 43.5   & 42.4     & 34.0    & 30.6  &  35.6  \\
LLaVA-OneVision-7B~\cite{li2024llavaov}  & 47.7  & 20.2   & 47.4      & 12.3    & 42.5   & 35.2     & 29.4    & 24.4  &  32.4  \\
\rowcolor{gray!15}
InternVL3-8B  & 68.1  & 39.0   & 48.4      & 33.6   & 48.3   &   36.4   & 27.3    & 35.4  &  42.1  \\

\rowcolor{gray!15}
InternVL3-38B &  71.7 & 50.2  & 46.1   & 41.7      & 53.5    & 38.6   & 28.9     & 60.7    & 48.9    \\
LLaVA-NeXT-Video-72B~\cite{zhang2024llavanext}   & 48.9  & 22.8   & 57.4      & 35.3    & 42.4   & 36.7     & 35.0    & 48.6  &  40.9  \\

LLaVA-OneVision-72B~\cite{li2024llavaov}  & 43.5  & 23.9   & 57.6      & 37.5    & 42.5   & 39.9     & 32.5    & 44.6  &  40.2  \\

\rowcolor{gray!15}
InternVL3-78B &  71.2 & 53.7  & 44.4   & 39.5      & 55.9    & 39.5   & 28.9     & 54.5    & 48.4    \\

\end{tabular}
}
\caption{\textbf{Performance of InternVL3 and other models on VSI-Bench.}
}
\label{tab:vis_benchmark}
\end{table*}

\subsection{Evaluation on Language Capability}

\begin{table*}[!t]
\centering
\small
\renewcommand{\arraystretch}{0.95}
\setlength\tabcolsep{5pt}
{\fontsize{8}{10}\selectfont 

\newcommand{\QwenOne}{\rotatebox{90}{\makecell{Qwen2.5-0.5B Chat}}}
\newcommand{\QwenTwo}{\rotatebox{90}{\makecell{Qwen2.5-1.5B Chat}}}
\newcommand{\QwenThree}{\rotatebox{90}{\makecell{Qwen2.5-7B Chat}}}
\newcommand{\QwenFour}{\rotatebox{90}{\makecell{Qwen2.5-14B Chat}}}
\newcommand{\QwenFive}{\rotatebox{90}{\makecell{Qwen2.5-32B Chat}}}
\newcommand{\QwenSix}{\rotatebox{90}{\makecell{Qwen2.5-72B Chat}}}

\newcommand{\InternVLThreeOne}{\rotatebox{90}{\makecell{InternVL3-1B}}}
\newcommand{\InternVLThreeTwo}{\rotatebox{90}{\makecell{InternVL3-2B}}}
\newcommand{\InternVLThreeThree}{\rotatebox{90}{\makecell{InternVL3-8B}}}
\newcommand{\InternVLThreeFour}{\rotatebox{90}{\makecell{InternVL3-9B}}}
\newcommand{\InternVLThreeFive}{\rotatebox{90}{\makecell{InternVL3-14B}}}
\newcommand{\InternVLThreeSix}{\rotatebox{90}{\makecell{InternVL3-38B}}}
\newcommand{\InternVLThreeSeven}{\rotatebox{90}{\makecell{InternVL3-78B}}}

\begin{tabular}{ll|rr|rr|rr|rr|rr|rr}
Dataset                         &  Version                                       & \QwenOne & \InternVLThreeOne & \QwenTwo & \InternVLThreeTwo & \QwenThree & \InternVLThreeThree & \QwenFour & \InternVLThreeFive & \QwenFive & \InternVLThreeSix & \QwenSix & \InternVLThreeSeven \\
\midrule
MMLU             & 4d595a & 46.4         & 49.8         & 61.8         & 64.8         & 74.2       & 77.3         & 79.5        & 82.1          & 83.3        & 85.4          & 84.4        & 86.9          \\
CMMLU            & c13365 & 47.2         & 56.7         & 62.9         & 72.2         & 78.8       & 84.4         & 82.6        & 85.8          & 85.8        & 88.7          & 87.4        & 89.9          \\
C-Eval           & 2daf24 & 53.5         & 59.0         & 66.2         & 73.3         & 77.8       & 84.5         & 81.4        & 85.6          & 86.5        & 89.2          & 88.1        & 89.5          \\
GAOKAO           & 4c31db & 30.9         & 46.6         & 53.7         & 67.7         & 81.3       & 89.5         & 86.9        & 91.2          & 90.8        & 93.5          & 91.0        & 93.1          \\
\midrule
TriviaQA         & 2121ce & 24.2         & 21.5         & 39.8         & 41.2         & 55.8       & 51.5         & 65.1        & 67.4          & 65.8        & 70.1          & 74.0        & 74.7          \\
NaturalQuestions & 3dcea1 & 8.2          & 8.5          & 15.2         & 15.9         & 17.9       & 28.2         & 19.7        & 31.4          & 19.7        & 31.0          & 23.8        & 39.0          \\
C3               & 8c358f & 35.2         & 66.3         & 81.2         & 84.7         & 90.8       & 95.1         & 92.1        & 96.3          & 92.3        & 97.4          & 96.1        & 97.6          \\
RACE-High        & 69ee4f & 51.5         & 68.8         & 76.0         & 84.6         & 86.8       & 90.8         & 89.6        & 93.0          & 91.5        & 94.2          & 91.7        & 94.2          \\
\midrule
WinoGrande       & b36770 & 47.2         & 52.9         & 56.5         & 61.9         & 71.5       & 78.1         & 79.1        & 84.3          & 83.8        & 86.7          & 83.9        & 87.8          \\
HellaSwag        & e42710 & 39.3         & 47.0         & 62.0         & 73.8         & 85.4       & 90.2         & 90.5        & 93.0          & 92.1        & 95.5          & 92.7        & 95.6          \\
BBH              & 5b92b0 & 21.5         & 34.5         & 39.7         & 52.0         & 65.7       & 77.4         & 73.0        & 82.5          & 85.5        & 87.7          & 85.4        & 85.2          \\
\midrule
GSM8K            & 1d7fe4 & 39.0         & 47.2         & 61.6         & 72.5         & 80.1       & 83.1         & 82.4        & 88.4          & 84.7        & 89.7          & 88.2        & 90.5          \\
MATH             & 393424 & 27.8         & 32.7         & 49.3         & 57.3         & 72.6       & 72.2         & 73.7        & 76.3          & 81.1        & 72.2          & 81.4        & 78.9          \\
TheoremQA        & 6f0af8 & 12.3         & 12.9         & 14.4         & 15.6         & 20.1       & 25.5         & 18.5        & 24.1          & 21.9        & 18.9          & 22.9        & 30.4          \\
\midrule
HumanEval        & 8e312c & 27.4         & 39.0         & 51.8         & 62.8         & 82.3       & 78.1         & 81.1        & 78.1          & 89.0        & 87.8          & 87.2        & 82.3          \\
MBPP             & a447ff & 38.5         & 47.5         & 51.4         & 60.7         & 74.3       & 69.3         & 76.7        & 75.1          & 83.7        & 77.4          & 86.8        & 76.7          \\
MBPP-CN          & 9114d5 & 19.6         & 30.6         & 34.4         & 45.8         & 64.4       & 64.4         & 75.4        & 67.2          & 77.8        & 75.4          & 76.0        & 76.0          \\
\midrule
Overall             & -           & 33.5         & 42.4         & 51.6         & 59.2         & 69.4       & 72.9         & 73.4        & 76.6          & 77.4        & 78.9          & 78.9        & 80.5         
\end{tabular}
}
\caption{\textbf{Comparison of language model performance across multiple benchmarks.} These results were obtained using the OpenCompass toolkit. We compare InternVL3 with  Qwen2.5 Chat models, whose corresponding pre-trained base models are employed as the initialization of the language component in InternVL3.  Please note that the evaluation scores of the Qwen2.5 series may differ from those officially reported, as we have adopted the prompt versions provided in the table across all datasets for OpenCompass evaluation.}

\label{tab:language_model_comparison}
\end{table*}

\begin{table*}[t!]
\scriptsize
\centering
{\fontsize{8}{10}\selectfont 
\renewcommand{\arraystretch}{0.95}
\setlength\tabcolsep{2.6pt}
\begin{tabular}{c|lccccccccccccc}
\toprule
\multirow{2}{*}{V2PE}       & \multirow{2}{*}{\( \delta \)} & TextVQA        & VizWiz        & ChartQA       & DocVQA        & AI2D          & InfoVQA       & GQA           & SQA-I         & \multirow{2}{*}{POPE} & Tiny           & MMMU          & SEED v1       & \multirow{2}{*}{Overall} \\
                            &                               & val            & val           & test avg      & val           & test          & val           & test          & test          &                       & LVLM           & val           & image         &         \\

\midrule
\no               & --                            & 78.4           & 61.7          & 81.4          & 89.4          & 81.1          & 69.4          & 60.8          & 94.4          & 87.9                  & 348.5          & 52.6          & 75.6          & 75.2    \\
\midrule
\multirow{5}{*}{\yes} & 1/256                         & 78.0           & 61.7          & 81.2          & 88.5          & 81.0          & 67.7          & 61.0          & 94.4          & 88.3                  & 345.3          & 52.9          & 75.9          & 75.0    \\
                            & 1/64                          & 78.3           & 62.0          & 81.7          & 89.4          & 81.3          & 69.6          & 60.9          & 94.7          & 88.3                  & 345.7          & 52.3          & 76.1          & 75.3    \\
                            & 1/16                          & 78.7           & 62.1          & 81.7          & 90.4          & 81.6          & 70.4          & 61.1          & \textbf{95.0} & 88.2                  & 345.0          & \textbf{53.3} & 76.1          & 75.6    \\
                            & 1/4                           & \textbf{ 79.0} & \textbf{62.2} & \textbf{82.4} & \textbf{91.0} & \textbf{81.8} & \textbf{71.7} & \textbf{61.2} & 94.9          & 88.1                  & 345.8          & 52.6          & \textbf{76.2} & 75.9    \\
                            & 1/1                           & 78.7           & 61.7          & 82.2          & 90.2          & 81.7          & 71.4          & 61.2          & 94.6          & \textbf{ 88.5}        & \textbf{347.2} & 52.4          & 76.1          & 75.7    \\ 
\bottomrule
\end{tabular}}
\caption{\textbf{Performance of the pre-trained InternVL3-8B model on multimodal benchmarks with different positional encoding strategies.} When employing V2PE, the impact of different positional increment values \( \delta \) is systematically evaluated.
}
\label{v2pe_eff}
\end{table*}

Table~\ref{tab:language_model_comparison} presents the performance evaluation of language capabilities across a diverse array of benchmarks. These benchmarks cover comprehensive assessments in general knowledge, linguistic understanding, reasoning, mathematics, and coding tasks, such as MMLU~\cite{hendrycks2020measuring}, CMMLU~\cite{li2023cmmlu}, C-Eval~\cite{huang2023ceval}, GAOKAO-Bench~\cite{Zhang2023gaokao}, TriviaQA~\cite{joshi2017triviaqa}, NaturalQuestions~\cite{naturalquestion,sun2019investigating}, RACE~\cite{lai2017race}, WinoGrande~\cite{sakaguchi2020winogrande}, HellaSwag~\cite{zellers2019hellaswag}, BigBench Hard~\cite{suzgun2023bigbench}, GSM8K-Test~\cite{cobbe2021training}, MATH~\cite{DBLP:conf/nips/HendrycksBKABTS21}, TheoremQA~\cite{DBLP:conf/emnlp/ChenYKLWMXWX23}, HumanEval~\cite{chen2021evaluating}, MBPP~\cite{austin2021program}, and MBPP-CN~\cite{austin2021program}.

In particular, the experiments conducted compare the performance of Qwen2.5 chat models against corresponding InternVL3 variants. Both model series share the same pre-trained Qwen2.5 base model as their initialization. After undergoing native multimodal pre-training followed by additional post-training, the InternVL3 series consistently demonstrates superior performance over the Qwen2.5 chat models across most evaluation benchmarks.

This observed enhancement in language capabilities primarily arises from several factors, including the integration of approximately 25\% pure-language data, joint parameter optimization during native multimodal pre-training, and the extensive use of high-quality textual corpora during the subsequent post-training stage. Such an approach not only strengthens multimodal comprehension but also significantly enhances language proficiency. Consequently, even when derived from identical pre-trained base models, the integrated multimodal and pure-text training strategy employed by InternVL3 results in substantially improved performance in language capabilities compared to the specialized training pipeline designed  for pure-text tasks used by the Qwen2.5 chat models.

\begin{figure}
    \centering
    \includegraphics[width=1\linewidth]{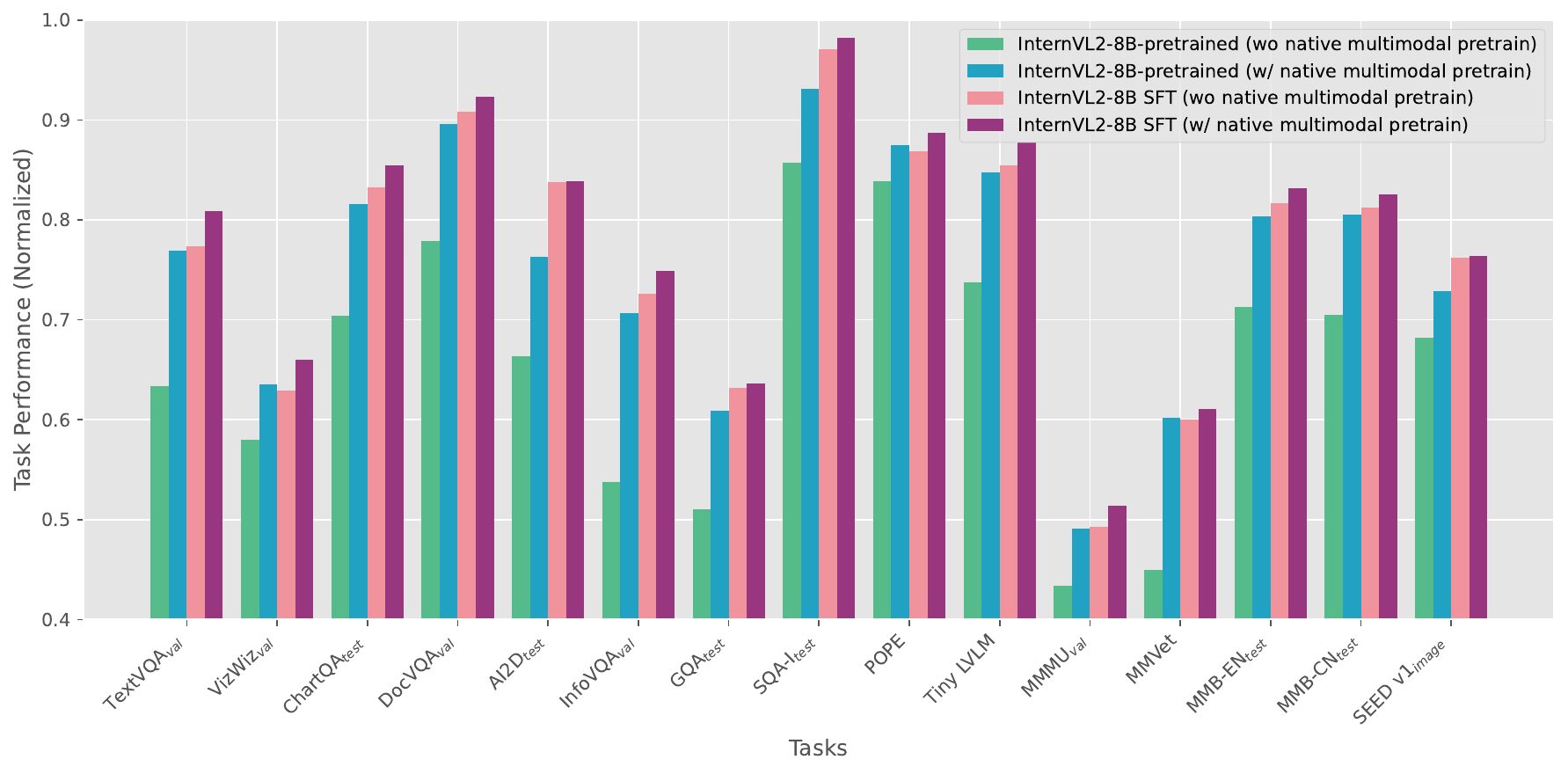}
    \caption{\textbf{Performance comparison on multimodal benchmarks under different training strategies.} Native multimodal pre-training endows MLLMs with strong multimodal capabilities, even without further post-training.
    }
    \label{fig:native_multimodal_performance}
\end{figure}
\begin{table*}[t!]

\centering
\small
\setlength\tabcolsep{3pt}
\begin{tabular}{l|c|ccccccc|l}
{Model}   & MPO    & {MMMU}        & {MathVista}        & {MathVision}        & {MathVerse}           & {DynaMath}        & {WeMath}        & {LogicVista}        & {Overall}    \\ 
\hline
\multirow{2}{*}{InternVL3-1B}   & \no   & 43.4          & 47.2               & 13.8                & 18.1                  & 4.2               & 14.7            & 31.1                & 24.6         \\
 &  \yes & 43.4          & 45.8               & 18.8                 & 18.7                   & 5.8             & 13.4            & 29.8             & 25.1 \gbf{+0.5}        \\

\hline
\multirow{2}{*}{InternVL3-2B}   & \no    & 49.1          & 59.0               & 22.0                & 23.2                  & 13.4              & 18.1            & 30.0                & 30.7         \\
 & \yes & 48.6           & 57.0               & 21.7                & 25.3                  & 14.6              & 22.4            & 36.9                & 32.4 \gbf{+1.7}        \\

\hline
\multirow{2}{*}{InternVL3-8B}   & \no       & 61.9          & 67.4               & 24.7                & 36.9                  & 22.8              & 32.7            & 43.2                & 41.4         \\
& \yes  & 62.7          & 71.6               & 29.3                & 39.8                  & 25.5              & 37.1            & 44.1                & 44.3 \gbf{+2.9}          \\
\hline
\multirow{2}{*}{InternVL3-9B}   & \no       & 59.0          & 68.8               & 28.9                & 32.2                  & 23.0              & 32.5            & 46.5                & 41.6         \\
& \yes  & 57.7          & 71.5               & 27.6                & 35.3                  & 26.7              & 33.8            & 49.2                & 43.1 \gbf{+1.5}          \\
\hline
\multirow{2}{*}{InternVL3-14B}   & \no       & 67.1          & 70.5               & 31.2                & 38.8                  & 27.9              & 38.1            & 49.9                & 46.2         \\
& \yes & 67.1          & 75.1               & 37.2                & 44.4                  & 31.3              & 43.0            & 51.2                & 49.9 \gbf{+3.7}          \\
\hline
\multirow{2}{*}{InternVL3-38B}   & \no       & 69.3          & 71.2               & 34.2                & 45.1                  & 22.2              & 41.7            & 54.4                & 48.3         \\
& \yes & 70.1          & 75.1               & 34.2                & 48.2                  & 35.3              & 48.6            & 58.4                & 52.8 \gbf{+4.5}         \\
\hline
\multirow{2}{*}{InternVL3-78B}   & \no       & 72.2          & 74.0               & 35.2                & 44.2                  & 31.7              & 42.5            & 53.5                & 50.5         \\ 
& \yes & 72.2          & 79.0               & 43.1                & 51.0                  & 35.1              & 46.1            & 55.9                & 54.6 \gbf{+4.1}         \\
\end{tabular}
\caption{{\textbf{Comparison of reasoning abilities before and after Mixed Preference Optimization (MPO).}
}
}
\label{tab:ablation_mpo}
\end{table*}

\subsection{Ablation Study}

\noindent \textbf{The Effectiveness of Native Multimodal Pre-Training.}
To assess the effectiveness of native multimodal pre-training, we conduct experiments on the InternVL2-8B model while keeping its architecture, initialization parameters, and training data entirely unchanged. Traditionally, InternVL2-8B employs a training pipeline that begins with an MLP warmup phase for multimodal alignment, followed by an instruction-tuning stage. In our experiments, we substitute the conventional MLP warmup phase with our native multimodal pre-training process. This modification isolates the contribution of native multimodal pre-training to the overall multimodal capability of the model.

The evaluation results in Figure~\ref{fig:native_multimodal_performance} show that the model with native multimodal pre-training exhibits performance on most benchmarks that is comparable to the fully multi-stage-trained InternVL2-8B baseline. Furthermore, when followed by instruction tuning on higher-quality data, the model demonstrates further performance gains across evaluated multimodal tasks. These findings underscore the efficiency of native multimodal pre-training in imparting powerful multimodal capabilities to MLLMs.

\noindent \textbf{The Evaluation of Variable Visual Position Encoding.}
To promote the multimodal capabilities in long-context scenarios, InternVL3 employs Variable Visual Position Encoding (V2PE) in its visual embedding. 
However, in the original V2PE~\cite{ge2024v2pe}, this specialized positional encoding for visual tokens did not yield benefits on multimodal tasks with moderate context lengths.
To further explore the efficacy of V2PE in a broader setting, we incorporated it during the native multimodal pre-training stage and evaluated the InternVL3-8B pre-trained model on standard multimodal benchmarks. 

As reported in Table~\ref{v2pe_eff}, the introduction of V2PE leads to significant performance gains across most evaluation metrics. In addition, our ablation studies—by varying the positional increment \( \delta \)—reveal that even for tasks primarily involving short contexts, relatively small \( \delta \) values can achieve optimal performance. These findings provide important insights for future efforts aimed at refining position encoding strategies for visual tokens in MLLMs.
It is important to note that, to ensure fair comparisons, all results elsewhere in this report maintain a fixed \( \delta = 1 \), except for the experimental results presented in Table~\ref{v2pe_eff}.

\noindent \textbf{Mixed Preference Optimization.}
Here, we demonstrate the effectiveness of MPO. As shown in Table~\ref{tab:ablation_mpo}, models fine-tuned with MPO demonstrate superior reasoning performance across seven multimodal reasoning benchmarks compared to their counterparts without MPO. Specifically, InternVL3-78B and InternVL3-38B outperform their counterparts by 4.1 and 4.5 points, respectively. Notably, the training data used for MPO is a subset of that used for SFT, indicating that the performance improvements primarily stem from the training algorithm rather than the training data.

\section{Conclusion}

\label{sec:conclusion}

We have introduced \textit{InternVL3}, a significant advancement in the InternVL series that implements a native multimodal pre-training paradigm. By jointly learning linguistic and multimodal capabilities during the pre-training phase, InternVL3 avoids the training complexities and optimization challenges typically associated with post-hoc MLLM training pipelines. Through the incorporation of variable visual position encoding (V2PE) for extended multimodal contexts, advanced post-training strategies—such as supervised fine-tuning and mixed preference optimization—and test-time scaling, InternVL3 establishes a new open-source benchmark across a wide range of multimodal tasks, while simultaneously preserving robust linguistic competencies.
Notably, \textit{InternVL3-78B} attains a 72.2-point score on the MMMU benchmark, exceeding previous open-source MLLMs and reducing the performance gap relative to leading proprietary counterparts (\eg, Gemini-2.5 Pro). In line with our commitment to fostering community-driven innovation in multimodal large language models, we will publicly release InternVL3’s training data and model weights, thereby encouraging further research and development in this rapidly evolving field.

{
    \small
    \bibliographystyle{plain}
    \bibliography{main}
}

\end{document}